\documentclass[AMS,STIX1COL]{WileyNJD-v2}
\usepackage{subfigure}

\articletype{Article Type}%

\begin{document}
	
	\title{Partially Observable Mean Field Multi-Agent Reinforcement Learning Based on Graph--Attention}

	\author[1]{Min Yang}
	\author[1]{Guanjun Liu*}
	\author[1]{Ziyuan Zhou}
	
	%\author[2]{Jiacun Wang}	
	
	%\authormark{AUTHOR ONE \textsc{et al}}

	\address[1]{\orgdiv{Department of Computer Science}, \orgname{Tongji University}, \orgaddress{\state{ Shanghai}, \country{China}}}
	%\address[2]{\orgdiv{Computer Science and Software Engineering Department}, \orgname{Monmouth University}, \orgaddress{\state{West Long Branch}, \country{NJ
			%	07764 USA}}}
	
	%\address[2]{\orgdiv{Org Division}, \orgname{Org Name}, \orgaddress{\state{State name}, \country{Country name}}}
	%
	%\address[3]{\orgdiv{Org Division}, \orgname{Org Name}, \orgaddress{\state{State name}, \country{Country name}}}
	
	%\corres{Guanjun Liu, Department of Computer Science, Tongji University, Shanghai,201800,China. \email{liuguanjuan@tongji.edu.cn}}

	\abstract[Summary]{Traditional multi-agent reinforcement learning algorithms are difficultly applied in a large-scale multi-agent environment. The introduction of mean field theory has enhanced the scalability of multi-agent reinforcement learning in recent years. This paper considers partially observable multi-agent reinforcement learning (MARL), where each agent can only observe other agents within a fixed range. This partial observability affects the agent's ability to assess the quality of the actions of surrounding agents. This paper focuses on developing a method to capture more effective information from local observations in order to select more effective actions. Previous work in this field employs probability distributions or weighted mean field to update the average actions of neighborhood agents, but it does not fully consider the feature information of surrounding neighbors and leads to a local optimum. In this paper, we propose a novel multi-agent reinforcement learning algorithm, Partially Observable Mean Field Multi-Agent Reinforcement Learning based on Graph--Attention (GAMFQ) to remedy this flaw. GAMFQ uses a graph attention module and a mean field module to describe how an agent is influenced by the actions of other agents at each time step. This graph attention module consists of a graph attention encoder and a differentiable attention mechanism, and this mechanism outputs a dynamic graph to represent the effectiveness of neighborhood agents against central agents. The mean--field module approximates the effect of a neighborhood agent on a central agent as the average effect of effective neighborhood agents. We evaluate GAMFQ on three challenging tasks in the MAgents framework. Experiments show that GAMFQ outperforms baselines  including the state-of-the-art partially observable mean-field reinforcement learning algorithms. The code for this paper is here \url{https://github.com/yangmin32/GPMF}.}
	

			\jnlcitation{\cname{
					\author{Min Yang}, 
					\author{Guanjun Liu}, 
					\author{Ziyuan Zhou}, 
					\author{Jiacun Wang}} (\cyear{2023}), 
				\ctitle{Partially Observable Mean Field Multi-Agent Reinforcement Learning Based on Graph Attention Network for UAV Swarms}, \cjournal{Drones}, \cvol{2023;7(7):476}. \doi{10.3390/drones7070476}\url{https://www.mdpi.com/2504-446X/7/7/476} }
			
	\maketitle
	
	%\footnotetext{\textbf{Abbreviations:} ANA, anti-nuclear antibodies; APC, antigen-presenting cells; IRF, interferon regulatory factor}
	
	\section{Introduction}	
	Reinforcement learning has been widely used in video games \cite{vinyals2019grandmaster} and recently in education \cite{gu2023metaverse}. For multi-agent reinforcement learning (MARL) \cite{zhang2021multi}, it involves multiple autonomous agents that make autonomous decisions to accomplish some specific competitive or cooperative tasks by maximizing global reward, it has been applied in some real-world scenarios such as autonomous mobile \cite{schmidt2022introduction} drone swarm confrontation\cite{azar2021drone} and multi-UAV collaboratively delivering goods \cite{shi2022marl}. For example, in some of the drone swarm adversarial tasks, drones need to make actions based on autonomous decisions. Due to the inevitable death of some drones in the confrontation environment \cite{zhou2022romfac}, the surviving drones must constantly evolve their strategies in real-time during the interaction with the environment to obtain the overall maximum reward. In order to make better interaction among agents, it is required that each agent in the multi-agent system can effectively perceive environmental information and fully acquire the information of surrounding agents. 
	
	However, the global communication cost among multiple agents is high, and in many practical tasks, each agent only observes part of the environmental information.  {Take the task of Autonomous Driving as an example, each vehicle makes decision in the limited observation space which is a typical local observation scene. Each agent can only rely on limited observation information in the local observation environment,} therefore the agent needs to learn a decentralized strategy. There are two common decentralization strategies. One is Centralized Training and Decentralized Execution (CTDE), which requires agents to communicate with each other during training and to independently make decisions based on their own observations during testing in order to adapt to large-scale multi-agent environments. Some classic algorithms using the CTDE framework such as MADDPG \cite{lowe2017multi}, QMIX \cite{rashid2018qmix} and MAVEN \cite{mahajan2019maven}. Another one takes the policy of decentralized training and decentralized execution, in which each agent can only observe part of the information during the training and testing phases, which is closer to the real environment with limited communication. Especially large-scale multi-agent environments are complex and non-stationary \cite{hernandez2019survey}, it is difficult for agents to observe the entire environment globally, limiting their ability to find the best actions. Furthermore, as the number of agents increases, joint optimization of all information in a multi-agent environment may result in a huge joint state-action space, which also brings scalability challenges. This paper focuses on the second strategy.
	
	Traditional multi-agent reinforcement learning algorithms are difficult to be applied in large-scale multi-agent environments, especially when the number of agents is exponential. Recent studies address the scalability issues of multi-agent reinforcement learning \cite{yang2018mean,xie2021learning,lauriere2022learning} by introducing mean-field theory, i.e., the multi-agent problem is reduced to a simple two-agent problem. However, Yang et al. \cite{yang2018mean} assumes that each agent can observe global information, which is difficult to apply in some real tasks. Therefore, it is necessary to study large-scale multi-agent reinforcement learning algorithms in partially observable cases  {\cite{cai2022reinforcement}}. In addition, researchers have intensively studied mean-field-based multi-agent reinforcement learning algorithms to improve performance  {in partially observable cases}. One way is to further decompose the Q-function of the mean field-based multi-agent reinforcement learning algorithm \cite{ZhangY0XL21,gu2021mean}. Another way uses probability distribution or weighted mean field to update the mean action of neighborhood agents \cite{fang2020large,zhou2020multi,Srirampomfrl2021,wu2022weighted}. 
	Hao \cite{haovery} combined the graph attention with the mean field to calculate the interaction strength between agents when agents interact, but only considered the scene where the agent has a fixed relative position, and the agents can observe the global information. The difference is that when the agent is partially observable, we consider the dynamic change of the agent's position and the death scene of the agent, and construct a more flexible partial observable graph attention network based on the mean field.
	
	However, for partially observable multi-agent mean field reinforcement learning, the existing methods do not fully consider the feature information of the surrounding neighbors, which will lead to falling into local optimum. {This paper focuses on identifying the neighborhood agents that may have the greater influence on the central agent in a limited observation space,} in order to avoid the local optimum issue. Since the graph neural network \cite{wu2020comprehensive} can fully aggregate the relationship between the central agent and its surrounding neighbors, we propose a graph attention-based mechanism to calculate the importance of neighbor agents to estimate the average action more efficiently.
	
	The main contributions of this paper are as follows:
	\begin{itemize}
		\item We propose a partially observable mean--field reinforcement learning based on the graph--attention (GAMFQ), which can learn a decentralized agent policy from an environment without requiring global information of an environment. In the case of partially observable large-scale agents, the judgment of the importance of neighbor agents is insufficient in our GAMFQ.
		\item We theoretically demonstrate that the settings of the GAMFQ algorithm are close to Nash equilibrium.
		\item Experiments on three challenging tasks in the MAgents framework show that GAMFQ outperforms two baseline algorithms as well as the state-of-the-art partially observable mean-field reinforcement learning algorithms.
	\end{itemize}
	
	\section{Related work}
	
	Most of the recent MARL algorithms for partial observability research are model-free reinforcement learning algorithms based on the CTDE framework. The most classic algorithm MADDPG \cite{lowe2017multi} introduces critics that can observe global information in training to guide actor training, but only use actors with local observation information to take actions in testing. QMIX\cite{rashid2018qmix} uses a hybrid network to combine the local value functions of a single agent, and adds global state information assistance in the training and learning process to improve the performance of the algorithm. MAVEN \cite{mahajan2019maven} is able to solve complex multi-agent tasks by introducing latent spaces for hierarchical control by value-mixing and policy-based approaches. However, these multi-agent reinforcement learning algorithm using the CTDE framework is difficult to scale to large-scale multi-agent environments, because there will be hard-to-observe global information that prevents the agents from training better policies.
	
	For large-scale multi-agent environments, Yang et al. \cite{yang2018mean} introduced the mean--field theory, which approximates the interaction of many agents as the interaction between the central agent and the average effects from neighboring agents. However, partially observed multi-agent mean--field reinforcement learning algorithms still have a space to improve. Some researchers further decompose the Q-function of the mean field based multi-agent reinforcement learning algorithm. Zhang et al. \cite{ZhangY0XL21} trained agents through the CTDE paradigm, transforming each agent's Q-function into its local Q-function and its mean field Q-function, but this approach is not strictly partially observable. Gu et al. \cite{gu2021mean} proposes a mean field multi-agent reinforcement learning algorithm with local training and decentralized execution. The Q-function is decomposed by grouping the observable neighbor states of each agent in a multi-agent system, so that the Q-function can be updated locally. In addition, some researchers have focused on improving the mean action in mean field reinforcement learning. Fang et al. \cite{fang2020large} adds the idea of mean field to MADDPG, and proposes a multi-agent reinforcement learning algorithm based on weighted mean field, so that MADDPG can adapt to large-scale multi-agent environment.  {Wang et al. \cite{wu2022weighted} propose a weighted mean-field multi-agent reinforcement learning algorithm based on reward attribution decomposition by approximating the weighted mean field as a joint optimization of implicit reward distribution between a central agent and its neighbors.} Zhou et al. \cite{zhou2020multi} uses the average action of neighbor agents as a label, and trained a mean field prediction network to replace the average action. Subramanian et al. \cite{Srirampomfrl2021} proposed two multi-agent mean field reinforcement learning algorithms based on partially observable settings: POMFQ(FOR) and POMFQ(PDO), extracting partial samples from Dirichlet or Gamma distribution to estimate partial observable mean action. Although these methods achieve good results, they do not fully consider the feature information of surrounding neighbors.
	
	Graph Neural Networks (GNNs) are able to mine graph structures from data for learning. In multi-agent reinforcement learning, GNNs can be used to model interactions between agents. In recent work, graph attention mechanisms have been used for multi-agent reinforcement learning. Zhang et al. \cite{zhang2022h2gnn} integrated the importance of the information of surrounding agents based on the multi-head attention mechanism, effectively integrate the key information of the graph to represent the environment and improve the cooperation strategy of agents with the help of multi-agent reinforcement learning. DCG \cite{bohmer2020deep} decomposed the joint value function of all agents into gains between pairs of agents according to the coordination graph, which can flexibly balance the performance and generalization ability of agents. Li et al. \cite{LiGMAK21} proposed a deep implicit coordination graph (DICG) structure that can adapt to dynamic environments and learn implicit reasoning about joint actions or values through graph neural networks. Ruan et al. \cite{ruan2022gcs} proposed a graph-based coordination strategy, which decomposes the joint team strategy into a graph generator and a graph-based coordination strategy to realize the coordination behavior between agents. MAGIC \cite{niu2021multi} more accurately represented the interactions between agents during communication by modifying the standard graph attention network and compatible with differentiable directed graphs.
	
	In the dynamic MARL system where competition and confrontation coexist, it is very difficult to directly apply the graph neural network, because the agent will die, the graph structure of the constructed large-scale agent system has the problem of large spatial dimension. However, graph neural networks can better mine the relationship between features, and the introduction of mean-field theory can further improve the advantages of mean-field multi-agent reinforcement learning.
	
	Our approach differs from related work above in that it uses a graph attention mechanism to select surrounding agents that are more important to the central agent in a partially observable environment. GAMFQ uses a graph attention module and a mean field module to describe how an agent is influenced by the actions of other agents at each time step, where graph attention consists of a graph attention encoder and a differentiable attention mechanism, and finally outputs a dynamic graph to represent the effectiveness of the neighborhood agent to the central agent. The mean field module approximates the influence of a neighborhood agent on a central agent as the average influence of the effective neighborhood agents. Using these two modules together is able to efficiently estimate the average action of surrounding agents in partially observable situations. GAMFQ does not require global information about the environment to learn decentralized agent policies from the environment.
	
	\section{Motivation \& Preliminaries}
	{In this section, we represent discrete-time non-cooperative multi-agent task modeling as a stochastic game (SG). SG can be defined as a tuple $<{S}, {A}^{1}, \ldots, {A}^{N}, r^{1}, \ldots, r^{N}, p, \gamma>$, where $S$ represents the true state of the environment. Each agent $j \in \{1, \ldots, N\}$ chooses an action at each time step $a^{j} \in {A}^{j }$. The reward function for agent $j$ is $r^{j}: {S} \times {A}^{1} \times \cdots \times{A}^{N} \rightarrow { R}$. State transitions are dynamically represented as $p: {S} \times {A}^{1} \times \cdots \times {A}^{N} \rightarrow \Omega({S})$ . $\gamma$ is a constant representing the discount factor. It represents a stable state, and in this stable state, all agents will not deviate from the best strategy given to others. The disadvantage is that it cannot be applied to the coexistence of multiple agents. Yang et al. \cite{yang2018mean} introduced mean field theory, which approximates the interaction of many agents as the interaction between the average effect of a central agent and neighboring agents, and solves the scalability problem of SG.}
	
	{The Nash equilibrium of general and random games can be defined as a strategy tuple $\left(\pi_{*}^{1}, \cdots, \pi_{*}^{N}\right)$, for all $s \in S$ and $\forall \pi^{i} \in \Pi^{i}$, there is $v^{j}\left(s, \pi_{*}^{1}, \cdots, \pi_{*}^{i}, \cdots, \pi_{*}^{N}\right) \geq v^{j}\left(s, \pi_{*}^{1}, \cdots, \pi^{i}, \cdots, \pi_{*}^{N}\right)$. This shows that when all other agents are implementing their equilibrium strategy, no one agent will deviate from this equilibrium strategy and receive a strictly higher reward. When all agents follow the Nash equilibrium strategy, the Nash Q-function of agent $j$ is $Q_{*}^{j}(s, {a})$. Partially observable stochastic games can generate a partially observable Markov decision process (POMDP), we review the partially observable Markov decision (Dec-POMDP) in Section 3.1 and analyze the partially observable model from a theoretical perspective. Section 3.2 first introduces the globally observable mean-field multi-agent reinforcement learning, and then introduces the partially observable mean-field reinforcement learning algorithm (POMFQ) based on the POMDP framework, and analyzes the existing part of the observable in detail. The limitation of mean-field reinforcement learning POMFQ(FOR)\cite{Srirampomfrl2021} is that the feature information of surrounding neighbors is not fully considered. In a partially observable setting, each agent $j$ observable neighborhood agent information $o^j$ can be used to better mine the relationship between features through a graph attention network. Introducing graph attention networks into partially observable mean-field multi-agent reinforcement learning can further improve their performance, and Section 3.3 briefly introduces graph attention networks.}
	
	\subsection{Partially observable Markov decision process} 
	We mainly study partially observable Markov decisions (Dec-POMDP)  {\cite{cai2022reinforcement,oliehoek2016concise,zhang2021multi}}. The partially observable Markov decision process of $n$ agents can be represented as a tuple $\left\langle {N,S, \left\{{A^i} \right\}_{i = 1}^n,T,Z,R,O,\gamma } \right\rangle $, where $N = \{ 1, \ldots ,n\} $ represents the set of agents, $S$ represents the global state,  {$A^j$} represents the set of action spaces of the  {$j$-th} agent, $Z$ represents the observation space of the agents, and the agent  {$j$} receives observation  {${o^j}\in {O^j}$} through the observation function  {$Z(s,j):S \times N \to O$}, and the transition function $T:S \times {A^1} \times  \ldots  \times {A^n} \times S \mapsto [0,1]$ represents the environment transitions from a state to another one. At each time step $t$, the agent  {$j$} chooses an action  {${a_t^j} \in {A^j}$}, gets a reward  {${r^j_t}:S \times {A^j} \mapsto R$} w.r.t. a state and an action. $\gamma  \in [0,1]$ is a reward discount factor. Agent  {$j$} has a stochastic policy  {${\pi ^j}$} conditioned on its observation  {${o^j}$} or action observation history  {${\tau ^j} \in \left( {Z \times {A^j}} \right)$}, and according to the all agents's joint policy $\pi  \buildrel \Delta \over = \left[ {{\pi ^1}, \ldots ,{\pi ^N}} \right]$, The value function of agent $j$ under the joint strategy $\pi$ is the value function $v_{\pi}^{j}(s)=\sum_{t=0}^{\infty} \gamma^{t} {E}_{\pi, p}\left[r_{t}^{j} | s_{0}=s\right]$ can be obtained, and then the Q-function can be formalized as $Q_{\pi}^{j}(s, {a})=r^{j}(s,{a})+\gamma {E}_{s^{\prime} \sim p}\left[v_{\pi}^{j}\left(s^{\prime}\right)\right]$. Our work is based on the POMDP framework.

	\subsection{Partially Observable Mean Field Reinforcement Learning}
	Mean-field theory-based reinforcement learning algorithm \cite{yang2018mean} approximates interactions among multiple agents as two-agent interactions, where the second agent corresponds to the average effect of all other agents. Yang et al. \cite{yang2018mean} decomposes the multi-agent Q-function into pairwise interacting local Q-functions as follows:
	\begin{eqnarray}
		{Q^j_{\pi}}(s,a) = \frac{1}{{{N^j}}}\sum\limits_{k \in N(j)} {{Q^j_{\pi}}} \left( {s,{a^j},{a^k}} \right)
		\label{eq1}
	\end{eqnarray}
	where ${N^j}$ is the index set of the neighbors of the agent $j$ and ${a^j}$ represents the discrete action of the agent $j$ and is represented by one-shot coding. Mean field Q-function is cyclically updated according to Eq.\ref{eq2}-\ref{eq5}:
	\begin{eqnarray}
		{Q^j_{\pi}}\left( {{s_t},a_t^j,\bar a_t^j} \right)  = (1 - \alpha ){Q^j_{\pi}}\left( {{s_t},a_t^j,\bar a_t^j} \right) + \alpha \left[ {r_t^j + \gamma {v^j}\left( {{s_{t + 1}}} \right)} \right]
		\label{eq2}
	\end{eqnarray}
	where
	\begin{eqnarray}
		{v^j}\left( {{s_{t + 1}}} \right) \!=\! \sum\limits_{a_{t + 1}^j} {{\pi ^j}} \left( {a_{t + 1}^j\!\mid\! {s_{t + 1}},\tilde a_t^j} \right){Q^j_{\pi}}\left( {{s_{t + 1}},a_{t + 1}^j,\tilde a_t^j} \right)
		\label{eq3}
	\end{eqnarray}
	\begin{eqnarray}
		\bar a_t^j = \frac{1}{N}\sum\limits_{k \ne j} {a_t^k} ,a_t^k \sim {\pi ^k}\left( { \cdot \mid {s_t},\bar a_{t - 1}^k} \right)
		\label{eq4}
	\end{eqnarray}
	\begin{eqnarray}
		{\pi ^j}\left( {a_t^j\mid {s_t},\bar a_{t - 1}^j} \right) \!=\! \frac{{\exp \left( { - \beta {Q^j_{\pi}}\left( {{s_t},a_t^j,\bar a_{t - 1}^j} \right)} \right)}}{{\sum\limits_{a_t^{j'} \in {A^j}} {\exp } \left( { - \beta {Q^j_{\pi}}\left( {{s_t},a_t^{j'},\bar a_{t - 1}^j} \right)} \right)}}
		\label{eq5}
	\end{eqnarray}
	where $\bar a_t^j$ is the mean action of the neighborhood agent, $r_t^j$ is the reward for agent $j$ at time step $t$, ${v^j}$ is the value function of agent $j$, and $\beta $ is the Boltzmann parameter. Literature \cite{yang2018mean} assumes that each agent has global information, and for the central agent, the average action of the neighboring agents is updated by Eq. \ref{eq4}. However, in a partially observable multi-agent environment, the way of calculating the average action in Eq. \ref{eq4} is no longer applicable.

\begin{figure}[!h]
		\centerline{\includegraphics[width=6cm]{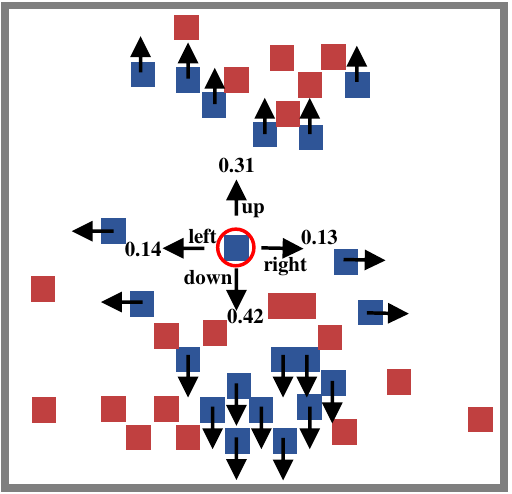}}
		\caption{A battle environment of the red and blue groups, where the red agent in the center is distributed by Dirichlet to calculate the action.\label{fig1}}
	\end{figure}

	In the case of partial observability, Subramanian et al. \cite{Srirampomfrl2021}  take $U$ samples from the Dirichlet distribution to update the average action of Eq. \ref{eq4}, and achieve better performance than the mean field reinforcement learning algorithm. The formula is as follows:
	\begin{eqnarray}
		\begin{array}{l}
			{D^j}(\theta ) \propto \theta _1^{{\eta _1} - 1 + {c_1}} \cdots \theta _L^{{\eta _L} - 1 + {c_L}}; \\
			\tilde a_{i,t}^j \sim {D^j}(\theta ;\eta  + c);\tilde a_t^j = \frac{1}{U}\sum\limits_{i = 1}^{i = U} {\tilde a_{i,t}^j} 
			\label{eq6}
		\end{array}
	\end{eqnarray}
	where $L$ denotes the size of the action space, ${c_1}, \ldots ,{c_L}$ denotes the number of occurrences of each action, $\eta $ is the Dirichlet parameter, $\theta $ is the classification distribution. But the premise of the Dirichlet distribution is to assume that the characteristics of each agent are independent to achieve better clustering based on the characteristics of neighboring agents. In fact, in many multi-agent environments, the characteristics of each agent has a certain correlation, but the Dirichlet distribution does not consider this correlation, which makes it unable to accurately describe the central agent and the neighborhood agents. There will be some deviations in the related information. Figure \ref{fig1} shows the process of a battle between the red and green teams, in which each agent can observe the information of the friendly agent, and the action space of the agent is $\{up, down, left, right\}$. The central agent enclosed by the red circle is affected by the surrounding friendly agents. We use the Dirichlet distribution to simulate and calculate the probability of the central agent moving in each direction, as shown below:
	\begin{eqnarray}
		\left\{ {\begin{array}{*{20}{l}}
				{{p_{up}} = 0.31}\\
				{{p_{down}} = 0.42}\\
				{{p_{left}} = 0.14}\\
				{{p_{right}} = 0.13}
		\end{array}} \right.
		\label{eq7}
	\end{eqnarray}
	
	It can be obtained that the probability of the agent moving down is the highest, which is essentially due to the large number of agents moving $down$. However, moving $up$ is the optimal action for the agent to form an encirclement trend with friends. The Dirichlet distribution results in a local optimal solution rather than finding the optimal action.
	
	Zhang et al. \cite{zhang2021coordination} believes that the correlation between two agents is crucial for multi-agent reinforcement learning. First, the paper calculates the correlation coefficient between each pair of agents, and then shields the communication among weakly correlated agents, thereby reducing the dimensionality of the state-action value network in the input space. Inspired by Zhang et al. \cite{zhang2021coordination}, for large-scale partially observable multi-agent environments, it is more necessary to select the importance of neighborhood agents. In our paper, we will adopt a graph attention method to filter out more important neighborhood agents, discard unimportant agent information, and achieve more accurate estimation of the average actions of neighborhood agents.
	
	\subsection{Graph Attention Network} 
	
	Graph neural network \cite{wu2020comprehensive} can better mine the graph structure form between data. Graph Attention Network (GAT) \cite{velivckovic2017graph} is composed of a group of graph attention layers, each graph attention layer acts on the node feature vector of node {$i$} denoting as {$m_{i}$} through a weight matrix {$W$} , and then uses softmax to normalize the neighbor nodes of the central node:
	\begin{eqnarray}
		{e_{ij}} =(W m_{i} \| W m_{j}) 
	\end{eqnarray}
	\begin{eqnarray}
		{\alpha _{ij}} = {\mathop{\rm softmax}\nolimits} j\left( {{e_{ij}}} \right) = \frac{{\exp \left( {{e_{ij}}} \right)}}{{\sum\limits_{k \in N_j} {\exp } \left( {{e_{jk}}} \right)}}
		\label{eq8}
	\end{eqnarray}
	where ${e_{ij}}$ is the attention coefficient of each node, indicating the importance of node  {$i$} to node  {$j$}. Finally, the output features are obtained by weighting the input features  {${h_i}$}, and the update rule for each node $j$ is:
	\begin{eqnarray}
		{{e_j} = \sigma \left( {\sum\limits_{i \in {N_j}} {\alpha _{ij}W{h_i}} } \right)}
		\label{eq9}
	\end{eqnarray}
	where  {${e_j}$} represents the feature of node  {$j$},  {${N_j}$} is the set of adjacent nodes of node  {$j$}, and $\sigma ( \cdot )$ is a nonlinear activation function.
	
	\section{Approach}
\begin{figure}[!h]
		\centerline{\includegraphics[width=0.96\textwidth]{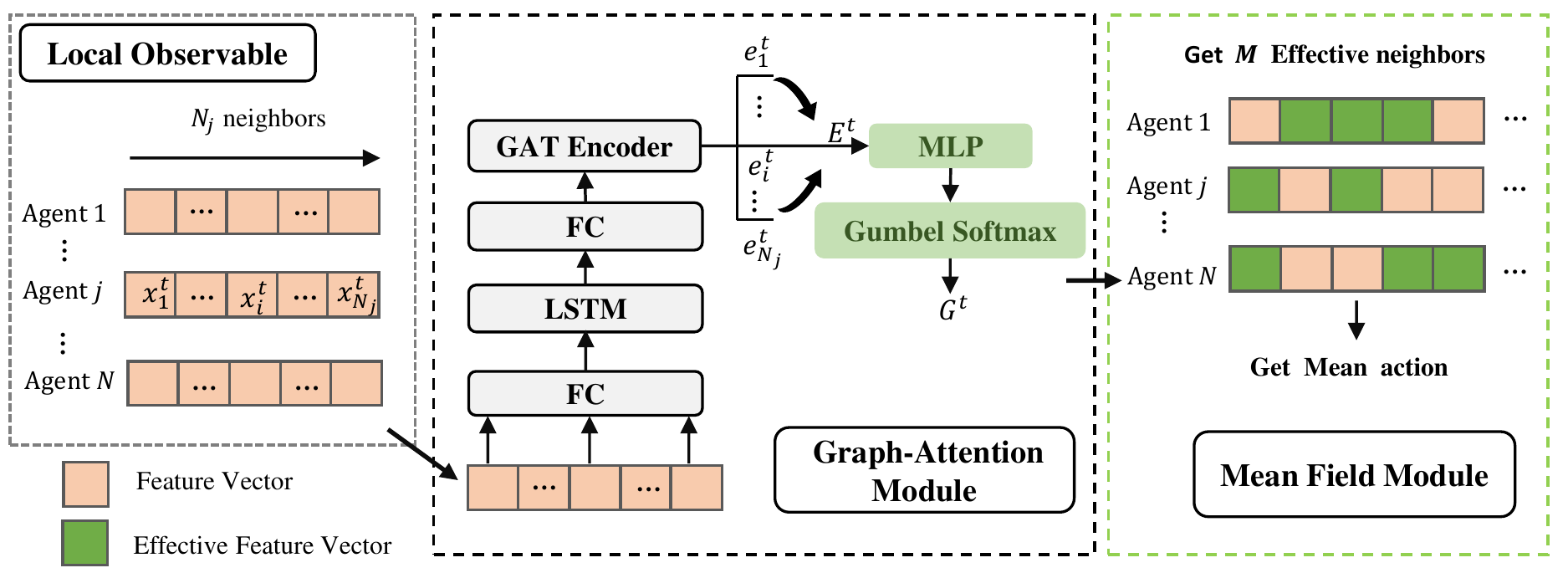}}
		\caption{Schematic of GAMFQ. Each agent can observe the feature information of other agents within a fixed range, input it into the Graph--Attention Module, and output an adjacency matrix to represent the effectiveness of the neighborhood agent to the central agent.\label{fig2}}
	\end{figure}

	{In this section, we propose a novel method called Partially Observable Mean Field Multi-Agent Reinforcement Learning based on Graph--Attention (GAMFQ), which can be applied to large-scale partially observable MARL tasks, where the observation range of each agent is limited, and the feature information of other agents in the fixed neighborhood is intelligently observed. The overall architecture of the GAMFQ algorithm is depicted in Figure \ref{fig2}, including two important components: the Graph Attention Module and the Mean Field Module: (i) In our Graph--Attention Module, the information observed locally by each agent is spliced firstly. Then the high-dimensional feature representations are obtained by a latent space mapping process which followed by a one-layer LSTM network to obtain the time-series correlation of the target agent, and the hidden layer of the LSTM is used as the input of the graph attention module to initialize the constructed graph nodes. Then to enhance the aggregation of neighbor agents to target agent, a similar process is implemented as a FC mapping network followed by a GAT layer. After that, the final representation of agents are obtained by a MLP layer with the input of the representations of target agent and other observable agents. Finally, we adopt layer-normalized method to obtain the adjacency matrix $\left\{ {G^t} \right\}_1^N$ via Gumbel Softmax. (ii) The Mean Field Module utilizes the adjacency matrix $\left\{ {G^t} \right\}_1^N$ from Graph Attention Module to obtain adopting action from important neighbor agents, in which the joint Q-function of each agent $j$ approximates the Mean-Field Q-function ${Q^j}(s,a) \approx Q_{{\rm{POMF}}}^j\left( {s,{a^j},{{\tilde a}^j}} \right)$ of important neighbor agents, where the Q-value is partially observable mean-field(POMF) Q-value, and $\tilde a^j$ is the average action of the important neighborhood a gents that is partially observable by agent $j$. Each component is described in detail below.}

	\subsection{Graph--Attention Module}
	To more accurately re-determine the influence of agent $j$'s neighbor  {${N_j}$} on itself, we need to be able to extract useful information from the local observations of agent $j$. The local observations of each agent include the embedding information of neighboring agents. For each agent $j$ and each time step $t$, the information of a local observation of length $L_j$, is expressed as $o_j^t=\left( {x_1^{t},x_2^ {t}, \cdots ,x_{N_j}^{t}} \right)$, where $x_{N_j}^{t}$ represents the feature of the ${N_j}$-th neighbor agent of agent $j$,  { and $o_j^t \in R^ {{N_j} \times D}$, $x_i^{t} \in R^ {1 \times D}$.} $L_j$ is concatenated from the embedding features of each neighbor. Our goal is to learn an adjacency matrix $\left\{ {G^t} \right\}_1^N$ to extract more important embedding information for the agent $j$ from local observations at each time step $t$. Since graph neural networks can better mine the information of neighbor nodes, we propose a graph attention structure suitable for large-scale multi-agent systems. This structure focuses on information from different agents by associating weights to observations based on the relative importance of other agents in their local observations. The Graph--Attention structure is constructed by concatenating a graph attention encoder and a differentiable attention mechanism. For the local observation $o_j^t$ of agent $j$ at time step $t$, $o_j^{t^{\prime}}$ is first encoded using a fully connected layer (FC) , and is passed to the LSTM layerin order to generate the hidden state $h_j^t$ and cell state $c_j^t$ of agent $j$, where $h_j^t$ serves as the input of the graph attention module to initialize the constructed graph nodes:
	\begin{eqnarray}
		h_j^t,c_j^t = LSTM\left( {e\left( {o_j^{t}} \right),h_j^{t},c_j^{t}} \right)
		\label{eq10}
	\end{eqnarray}
	where $e( \cdot )$ is a fully connected layer representing the observed encoder.  {$h_j^t$} is encoded as a message:
	\begin{eqnarray}
		m_j^t = {e}\left( {h_j^t} \right)
		\label{eq11}
	\end{eqnarray}
	where $m_j^t$ is the aggregated information of the neighborhood agents observed by agent $j$ at time step $t$. The input encoding information ${M^t}$ is passed to the GAT encoder and hard attention mechanism, where the hard attention mechanism consists of MLP and Gumbel Softmax function. Finally, the output adjacency matrix $\left\{ {G^t} \right\}_1^N$ is used to determine which agents in the neighborhood have an influence on the current agent. The GAT encoder helps to efficiently encode the agent's local information, which is expressed as:
	\begin{eqnarray}
		\{ {M^t}\} _1^N= {f_{Sched}}\left( {m_1^t, \cdots ,m_N^t} \right)
		\label{eq12}
	\end{eqnarray}
	
	Additionally, we take the form of the same attention mechanism as GAT \cite{velivckovic2017graph}, expressed as:
	\begin{eqnarray}
		{\alpha _{ij}^S =\frac{{\exp \left( {Leaky{\mathop{\rm Re}\nolimits} LU\left( {a_S^T\left[ {{W_S}m_i^t||{W_S}m_j^t} \right]} \right)} \right)}}{{\sum\limits_{k \in N_j^t \cup\{j\}} {\exp } \left( {Leaky{\mathop{\rm Re}\nolimits} LU\left( {a_S^T\left[ {{W_S}m_j^t||{W_S}m_k^t} \right]} \right)} \right)}}}
		\label{eq13}
	\end{eqnarray}
	where $LeakyReLU( \cdot )$ is the activation function, ${a_S} \in {R^{D}}$ is the weight vector,  {$N_j^t\cup\{j\}$} represents the central agent  {$j$} and its observable neighborhood agent set, and ${W_S} \in {R^{D \times D}}$ is the weight matrix. The node feature of agent $j$ is expressed as:
	\begin{eqnarray}
		{e_j^t = ELU\left( {\sum\limits_{i \in N_j^t \cup j} {\alpha _{ij}^S} {W_S}m_i^t} \right)}
		\label{eq14}
	\end{eqnarray}
	where $ELU( \cdot )$ is an exponential linear unit function. Connecting the features of each node in pairs: $E_{i,j}^t = \left( {e_i^t||e_j^t} \right)$, we can get a matrix ${E^t} \in {R^{N \times N_j \times 2D }}$, where $E_{i,j}^t$ represents the relevant features of agent $j$. Taking ${E^t}$ as the input of MLP which is followed by a Gumbel Softmax function, the connected vector $G _j^t$ can be obtained. The connected vector $G _j^t$ consists of elements $g_{ij}$, where $i$ represents the neighbors of the central agent $j$. The element $g_{ij}^{t} = 1$ in the adjacency matrix indicates that the action of the agent $i$ will have an impact on the agent $j$. Conversely, $g_{ij}^{t} = 0$ means that the agent's actions have no effect on the agent $j$. 
	
	\subsection{Mean Field Module}
	This Graph-Attention method selects important $M_j$ agents from the neighbors  {${N_j}$} of agent $j$, and compute the average of the actions of the choosed neighbor agents:
	\begin{eqnarray}
		{\tilde a_t^j = \frac{1}{{{M_j}}}\sum\limits_{k \in {N_j}} {{a_t^k \cdot G_j^t}},\quad a_{t}^{k} \sim \pi^{k}\left(\cdot \mid s_{t}, \tilde{a}_{t}^{k}\right)} 
		\label{eq15}
	\end{eqnarray}
	where $\cdot$ is the element-wise multiplication.
	
	In the above formula, ${a^k}$ represents the important neighborhood agent for agent $j$. Then the Q--value of each agent is shown in Eq. \ref{eq16}. Note that the Q--value here is a partially observable Q--value. 
	\begin{eqnarray}
		{	{Q_{{\rm{GAMF}}}^j}\left( {s_t^j,a_t^j,\tilde a_t^j} \right) = (1 - \alpha ){Q_{{\rm{GAMF}}}^j}\left( {s_t^j,a_t^j,\tilde a_t^j} \right) + \alpha \left[ {r_t^j + \gamma v\left( {s_{t + 1}^j} \right)} \right]}
		\label{eq16}
	\end{eqnarray}
	where the value function $v^j$ is expressed as
	\begin{eqnarray}
		{{v^j}\left( {s_{t + 1}^j} \right)= \sum\limits_{a_{t + 1}^j} {{\pi ^j}} \left( {a_{t + 1}^j\mid s_{t + 1}^j,\tilde a_t^j} \right){Q_{{\rm{GAMF}}}^j}\left( {s_{t + 1}^j,a_{t + 1}^j,\tilde a_t^j} \right)}
		\label{eq17}
	\end{eqnarray}
	
	According to the above graph attention mechanism, more important neighborhood agents are obtained. The new average action $\tilde a_t^j$ is calculated by Eq.\ref{eq15}, and then the strategy $\pi _t^j$ of agent $j$ is updated by the following formula:
	\begin{eqnarray}
		{{\pi ^j}\left( {a_t^j\mid s_t^j,\tilde a_{t - 1}^j} \right) = \frac{{\exp \left( { - \beta {Q_{{\rm{GAMF}}}^j}\left( {s_t^j,a_t^j,\tilde a_{t - 1}^j} \right)} \right)}}{{\sum\limits_{a_t^{j^{\prime}} \in {A^j}} {\exp } \left( { - \beta {Q_{{\rm{GAMF}}}^j}\left( {s_t^j,a_t^{j^{\prime}},\tilde a_{t - 1}^j} \right)} \right)}}}
		\label{eq18}
	\end{eqnarray}
	
	\subsection{Theoretical Proof} 
This subsection is devoted to proving that the setting of GAMFQ is close to the Nash equilibrium. Subramanian et al. \cite{Srirampomfrl2021} showed that in partially observable cases, the fixed observation radius (FOR) setting is close to a Nash equilibrium, where the mean action of each agent's neighborhood agents is approximated by a dirichlet distribution. First, we state some assumptions, which are the same as literature\cite{Srirampomfrl2021}, and are followed by all the theorems and analyses below.
		\begin{assumption}
			For any $i$ and $j$, there is $\lim _{t \rightarrow \infty} \tau_{j}^{i}(t)=\infty$. $w.p.1$. 
			\label{assumption1}
		\end{assumption}
		This assumption guarantees a probability of 1 that old information is eventually discarded.
		\begin{assumption}
			Suppose some measurability conditions are as follow: 
			(1) $x(0)$ is $\mathcal{F}(0)$-measurable.
			(2) For each $i$,$j$ and $t$, $w_{i}(t)$ is $\mathcal{F}(t+1)$-measurable.
			(3) For each $i, j$ and $t$, $\alpha_{i}(t)$ and $\tau_{j}^{i}(t)$ are $\mathcal{F}(t)$-measurable.
			(4) For each $i$ and $t$, satisfy $\mathrm{B}\left[w_{i}(t) | \mathcal{F}(t)\right]=0$.
			(5) $\mathrm{B}\left[w_{i}^{2}(t) | \mathcal{F}(t)\right] \leq A+B \max _{j} \max _{\tau \leq t}\left|x_{j}(\tau)\right|^{2}$, where $A$ and $B$ are deterministic constants.
			\label{assumption2}
		\end{assumption}
		\begin{assumption}
			The learning rates satisfy $0 \leq \alpha_{i}(t)<1$.
			\label{assumption3}
		\end{assumption}
		\begin{assumption}
			Suppose some conditions for the $F$ mapping are as follows:
			(1) If $x \leq y$, then $F(x) \leq F(y)$, that is, $F$ is monotonic;
			(2) $F$ is continuous;
			(3) When $t \rightarrow \infty$, $F$ is limited to the interval $\left[x^{*}-D, x^{*}+D\right]$, where $x^{*} $ is some arbitrary point;
			(4) If $e \in \mathcal{R}^{n}$ is a vector that satisfies all components equal to 1, then $F(x)-p e \leq F(x+p e) \leq F(x+p e) +p e$, where $p$ is a positive scalar.
			\label{assumption4}
		\end{assumption}
		\begin{assumption}
			Each action-value pair can be accessed indefinitely, and the reward is limited.
			\label{assumption5}
		\end{assumption}
		\begin{assumption}
			Under the limit $t \rightarrow \infty$ of infinite exploration, the agent's policy is greedy. 
			\label{assumption6}
		\end{assumption}
		This assumption ensures that the agent is rational.
		\begin{assumption}
			In each stage of a stochastic game, a Nash equilibrium can be regarded as a global optimum or saddle point.
			\label{assumption7}
		\end{assumption}
        Based on these assumptions, Subramanian et al. \cite{Srirampomfrl2021} give the following lemma.
        \begin{lemma}\cite{Srirampomfrl2021}
        	When the Q-function is updated using the partially observable update rule in Eq.\ref{eq2}, and assumptions \ref{assumption3}, \ref{assumption5}, and \ref{assumption7} hold, the following holds for $t \to \infty $:
        	\begin{eqnarray}
        		|{Q_*}({s_t},{a_t}) - {Q_{POMF}}({s_t},{a_t},{\tilde a_t})| \le 2D
        		\label{eq23}
        	\end{eqnarray}
        	where $Q_*$ is the Nash Q-value, $Q_{POMF}$ is the partially observable mean-field Q-function, and $D$ is the bound of the $F$ map. The probability that the above formula holds is at least $\delta^{L-1}$ , where $L=|A|$. \label{theorem4}
        \end{lemma}
In our GAMFQ setting, for partially observable neighborhood agents, we choose to select a limited number of important agents by using graph attention, and then update the POMF Q function. The following theorem proves that the setting of GAMFQ is close to Nash equilibrium.
    
		\begin{theorem}
			The distance between the MFQ (globally observable) mean action $\bar{a}$ and the GAMFQ (partially observable) mean action $\tilde{a}$ satisfies the following formula:
			\begin{eqnarray}
				{\left|\tilde{a}_t^j-\bar{a}_t^j\right| \leq \sqrt{\frac{1}{2 N_j} \log \frac{2}{\delta}}}
			\end{eqnarray}
			When $t \rightarrow \infty$, the probability $>=\delta$, where $N_j$ is the number of observed neighbor agents, $\tilde a$ is the partially observable mean action obtained by graph attention in Eq. \ref{eq15}, $\bar a$ is the globally observable mean action in Eq. \ref{eq4}.   \label{theorem1}
		\end{theorem}
				Assuming that each agent is globally observable, the mean of important agents selected by graph attention is close to the true underlying global observable $\bar a$. Since the GAMF Q-function is updated by taking finite samples through graph attention, the empirical mean is $\tilde a$.
		\begin{theorem}
			If the Q-function is Lipschitz continuous with respect to the mean action, i.e. $M$ is constant, then the MF Q-function $Q_{MF}$ and GAMF Q-function $Q_{GAMF}$ satisfy the following relation:
			\begin{eqnarray}
				\left|Q_{ {GAMF }}\left(s_{t}, a_{t}, \tilde{a}_{t-1}\right)-Q_{M F}\left(s_{t}, a_{t}, \bar{a}_{t-1}\right)\right| \leq M \times L \times \log \frac{2}{\delta} \times \frac{1}{2 N_j}
			\end{eqnarray}
			When the limit $t \rightarrow \infty$, the probability is $\geq(\delta)^{L-1}$, where $L=|A|$, $A$ is the action space of the agent. \label{theorem2}
		\end{theorem}
	In the proof of theorem \ref{theorem2}, first consider a Q-function that is Lipschitz continuous for all $\bar a$ and $\tilde{a}$.According to theorem \ref{theorem1}, the above formula can further deduce the result of theorem \ref{theorem2}. The total number of components is equal to the action space $L$. The bound of theorem \ref{theorem1} is probability $>=\delta$, and since there are $L$ random variables, the probability of theorem \ref{theorem2} is at least $(\delta)^{L-1}$. When the first $L-1$ random variable is fixed, the deterministic last $\bar a$ component satisfies the relationship that the sum of the individual components is 1. Since each agent's action is represented by a one-hot encoding, the $\tilde a^{\prime}$ component of GAMFQ also satisfies the relationship that the sum of the individual components is 1, and the component of the agent's average action does not change due to the application of graph attention. The proof of theorem \ref{theorem2} ends.
		\begin{theorem}
			A stochastic process in form $x_{i}(t+1)=x_{i}(t)+\alpha_{i}(t)\left(F_{i}\left(x^{i}(t)\right)-x_{i}(t)+w_{i}(t)\right)$ remains bounded in the range $[{x_*}-2D,{x_*}+2D]$ on limit $t \rightarrow \infty$ if assumptions \ref{assumption1},\ref{assumption2},\ref{assumption3} and \ref{assumption4} are satisfied, and are guaranteed not to diverge to infinity.Where $D$ is the boundary of the $F$ map in assumption \ref{assumption4}(4).
			\label{theorem3}
		\end{theorem}	
	This theorem can be proved in terms of Tsitsiklis\cite{tsitsiklis1994asynchronous} and by extension. The result of theorem \ref{theorem3} can then be used to derive theorem \ref{theorem4}.	
        \begin{theorem} 	
	When the Q-function is updated using the partially observable update rule in Eq.\ref{eq16}, and assumptions \ref{assumption3}, \ref{assumption5}, and \ref{assumption7} hold, the following holds for $t \to \infty $:
	\begin{eqnarray}
		|{Q_*}({s_t},{a_t}) - {Q_{GAMF}}({s_t},{a_t},{\tilde a_t})| \le 2D
		\label{eq23}
	\end{eqnarray}
	where $Q_*$ is the Nash Q-value, $Q_{GAMF}$ is the partially observable mean-field Q-function, and $D$ is the bound of the $F$ map. The probability that the above formula holds is at least $\delta^{L-1}$ , where $L=|A|$. \label{theorem4}
\end{theorem}
 Theorem \ref{theorem4} shows that the GAMFQ update is very close to the Nash equilibrium at the limit $t \rightarrow \infty$, i.e. reaching a plateau for stochastic policies. Therefore, the strategy of Eq.\ref{eq18} is approximately close to this plateau. Theorem \ref{theorem4} is an application of theorem \ref{theorem3}, using assumptions \ref{assumption3}, \ref{assumption5} and \ref{assumption7} .However, in MARL, reaching a Nash equilibrium is not optimal, but only a fixed-point guarantee. Therefore, to achieve better performance, each selfish agent will still tend to pick a limited number of samples. To balance theory and performance when selecting agents from the neighborhood, an appropriate number of agents (more efficient agents) need to be used for better multi-agent system performance. This paper uses the graph attention structure to filter out more important proxies, which can better approximate the Nash equilibrium.
	
	\subsection{Algorithm}
	
	The implementation of GAMFQ follows the related work of the previous POMFQ \cite{Srirampomfrl2021}, the difference is that the graph attention structure is used to select the neighborhood agents that are more important to the central agent when updating the average action. Algorithm \ref{alg1} gives the pseudocode of the GAMFQ algorithm. It obtains effective neighbor agents by continuously updating the adjacency matrix $G_{j}^{t}$ to update the agent's strategy.
	
	\begin{algorithm}
		\caption{Partially Observable Mean Field MARL Based on Graph--Attention}\label{alg1}
		\begin{algorithmic}
			\State Initialize the weights of Q-function $Q_{\phi^ j}$, $Q_{\phi^ j_{-}}$, replay buffer $B$, GAT encoder, MLP layers and mean action $\bar{a}^{j}$ for each agent $j \in 1, \ldots, N$.  \
			\For {$episode = 1, 2,\ldots, E$}
			\For {$t \le T$ and not terminal}
			\State For each agent $j$, calculate the hidden state $h_j^t$ according to Eq.\ref{eq10}, and encode $h_j^t$ as a message $m_j^t$ (Eq.\ref{eq11}).\
			\State For each agent $j$, sample $a^{j}$ fron policy induced by $Q_{\phi ^j} $(Eq.\ref{eq18}).\
			\State For each agent $j$, pass the encoded information $m_j^t$ to the GAT encoder and hard attention mechanism to output the adjacency matrix $G_{j}^{t}$.\
			\State For each agent $j$, calculate the new neighborhood agent mean action $\bar{a}^{j}$ by Eq.\ref{eq15}.\
			\State Receive the full state of environment $s_t$, action $a=\left[a^{1}, \ldots, a^{N}\right]$, reward $\left[r = r^{1}, \ldots, r^{N}\right]$, and the next state $s ^{\prime}=\left[ s ^{1}, \ldots, s ^{ N }\right]$.
			\State Store transition $\left\langle s, a, r, s^{\prime}, \bar{a}\right\rangle$ in $B$, where $\bar{ a }=\left[\bar{a}^{1}, \ldots, \bar{a}^{N}\right]$ is the mean action.\		
			\EndFor
			\For{$j = 1, \ldots, N$}
			\State Sample a minibatch of K experiences $\left\langle s, a, r, s^{\prime}, \bar{a}\right\rangle$ from replay buffer $B$.\
			\State Set $y^{j}=r^{j}+\gamma v_{\phi}\left(s^{\prime}\right)$ according to Eq.\ref{eq17}. \
			\State minimize the loss  $L\left(\phi^{j}\right)= \left(y^{j}-\right.\left.Q_{\psi_j}\left(s^{\prime}, a^{j}, \bar{a}^{j}\right)\right)^{2}$ to update Q network.\
			\EndFor
			\State For each agent $j$, update params of target network :$\phi^{j} \leftarrow \tau \phi^{j}+(1-\tau) \phi^{j}$.\
			\EndFor
		\end{algorithmic}
	\end{algorithm}
	
	\section{Experiments}
	In this section, we describe three different tasks based on the MAgent framework and give some experimental setup and training details for evaluating the GAMFQ performance.
	\subsection{Environments and Tasks}
	Subramanian et al. \cite{Srirampomfrl2021} designed three different cooperative-competitive strategies in the MAgent framework \cite{zheng2018magent} as experimental environments, and our experiments adopt the same environments. In these three tasks, the map size is set to 28*28, where the observation range of each agent is 6 units. The state space is the concatenation of the feature information of other agents within each agent's field of view, including location, health, and grouping information. The action space includes 13 move actions and 8 attack actions. In addition, each agent is required to handle at most 20 other agents that are closest. We will evaluate against these three tasks:
	
	\begin{itemize}
		\item \textbf{Multibattle environment:} There are two groups of agents fighting each other, each containing 25 agents. The agent gets -0.005 points for each move, -0.1 points for attacking an empty area, 200 points for killing an enemy agent, and 0.2 points for a successful attack. Each agent is 2*2 in size, has a maximum health of 10 units, and a speed of 2 units. After the battle, the team with the most surviving agents wins. If both teams have the same number of surviving agents, the team with the highest reward wins. The reward for each team is the sum of the rewards for the individual agents in the team.
		\item \textbf{Battle-Gathering environment:} There is a uniform distribution of food in the environment, each agent can observe the location of all the food. In addition to attacking the enemy to get rewards, each agent can also eat food to get rewards. Agents get 5 points for attacking enemy agents, and the rest of the reward settings are the same as the Multibattle environment.
		\item \textbf{Predator-Prey environment:} There are 40 predators and 20 prey, where each predator is a square grid of size 2*2 with a maximum health of 10 units and a speed of 2 units. Prey is a 1*1 square with a maximum health of 2 units and a speed of 2.5 units. To win the game, the predator must kill more prey, and the prey must find a way to escape. In addition, predators and prey have different reward functions, predators get -0.3 points for attacking space, 1 point for successfully attacking prey, 100 points for killing prey, -1 point for attacked prey, and 0.5 points for dying. Unlike the Multibattle environment, when the round ends for a fairer duel, if the two teams have the same number of surviving agents, it is judged as a draw.
	\end{itemize}
	
	\subsection{Evaluation}
	
	We consider four algorithms for the above three games: MFQ, MFAC \cite{yang2018mean}, POMFQ(FOR) and GAMFQ, where MFQ and MFAC are baselines and POMFQ(FOR) \cite{Srirampomfrl2021} is the state-of-the-art algorithm.
	
	The original baselines MFQ and MFAC were proposed by Yang et al. \cite{yang2018mean} based on global observability, and the idea was to approximate the influence of the neighborhood agents on the central agent as their average actions, thereby updating the actions of the neighborhood agents. We fix the observation radius of each agent in the baseline MFQ and MFAC and apply it to a partially observable environment, where neighbor agents are agents within a fixed range. The POMFQ(FOR) algorithm introduces noise in the mean action parameters to encourage exploration, uses Bayesian inference to update the Dirichlet distribution, and samples 100 samples from the Dirichlet distribution to estimate partially observable mean field actions. The GAMFQ algorithm judges the effectiveness of neighborhood agents within a fixed range through the graph attention mechanism, selects more important neighborhood agents, and updates the average action by averaging the actions of these agents.
	
	\subsection{Hyperparameters}
	In the three tasks, each algorithm was trained for 2000 epochs in the training phase, generating two sets of A and B sets of models. In the test phase, 1000 rounds of confrontation were conducted, of which the first 500 rounds were the first group A of the first algorithm against the second group B of the second algorithm, and the last 500 groups were the opposite. The hyperparameters of MFQ, MFAC, POMFQ(FOR) and GAMFQ are basically the same. Table \ref{tab1} lists the hyperparameters during training of the four algorithms, and the remaining parameters can be seen in  \cite{Srirampomfrl2021}.
	\begin{center}
		\begin{table}[!h]%
			\centering
			\caption{Hyperparameters for four algorithms training.\label{tab1}}%
			\begin{tabular*}{500pt}{@{\extracolsep\fill}lcccc@{\extracolsep\fill}}
				\toprule
				\textbf{Parameter} & \textbf{Value}  & \textbf{Description}  \\
				\midrule
				$\alpha$ & $10^{-4}$& learning rate \\
				$\beta$ & decays linearly from 1 to 0 & exploration rate\\
				$\gamma$ & 0.95 & discount rate \\
				$B$ & 1024 &  replay buffer \\
				$h$ & 64 & the hidden layer size in GAT  \\
				$K$ & 64 & mini-batch\\
				$temperature$ & 0.1 & the soft-max layer temperature of the actor in MFAC \\    
				\bottomrule
			\end{tabular*}
		\end{table}
	\end{center}
	
	\section{Results and discussion}
	
	In this section, we evaluate the performance of GMAFQ in three different environments, including Multibattle, Battle-Gathering, and Predator-Prey. We benchmark against two algorithms, MFQ and MFAC, and compare with the state-of-the-art POMFQ (FOR). We implement our method and comparative methods on three different tasks. Note that we only used 50 agents in our experiments and did not test more agents, this is because the proportion of other agents that each agent can see is more important than the absolute number. 
	\subsection{Reward}
\begin{figure}[htbp]
		\centerline{
			\subfigure[Train results of Multibattle game.]{
				\begin{minipage}[a]{0.3205\textwidth}
					\includegraphics[width=1\textwidth]{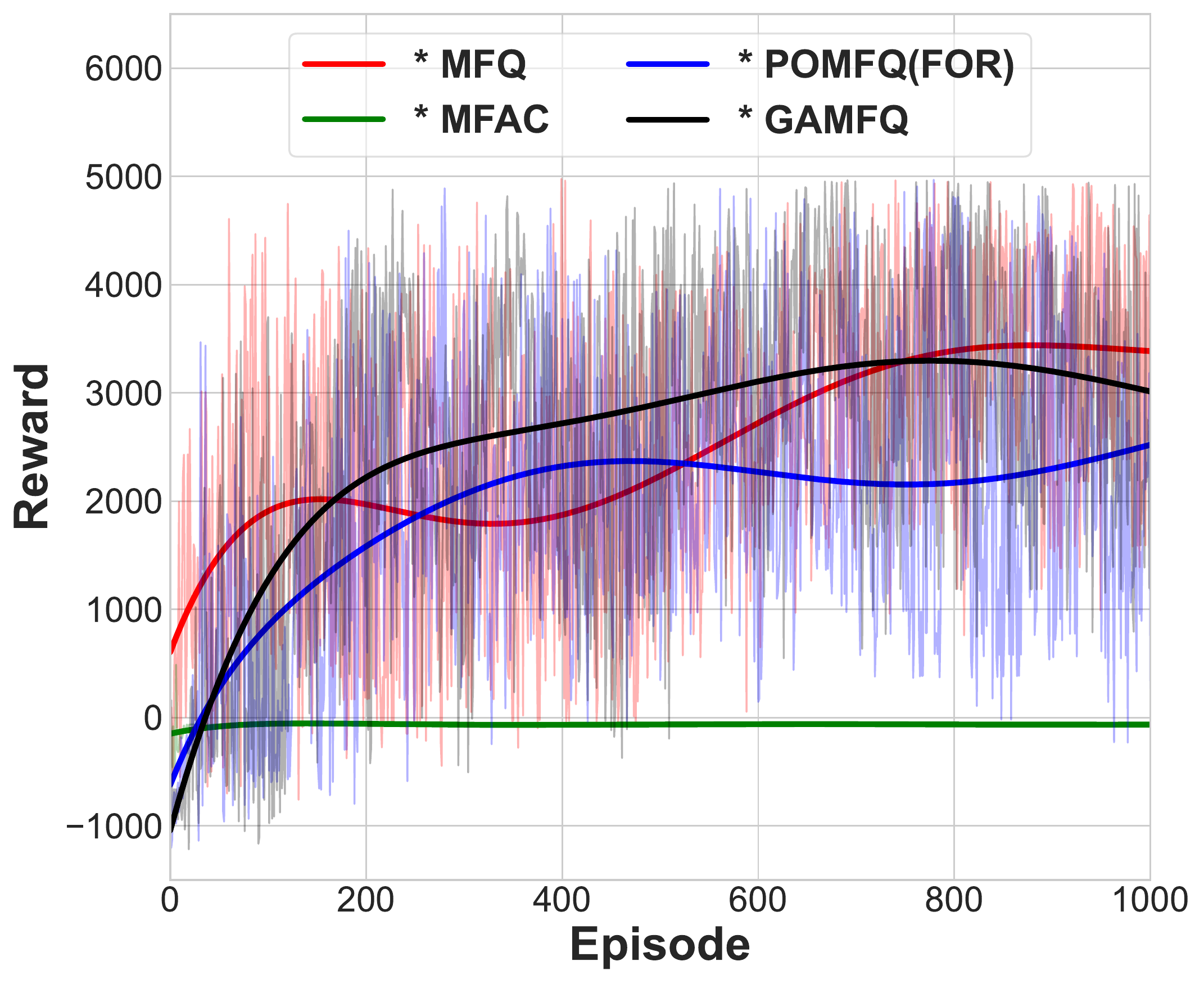}
				\end{minipage}
			}
			\subfigure[Train results of Battle-Gathering game.]{
				\begin{minipage}[a]{0.3205\textwidth}
					\includegraphics[width=1\textwidth]{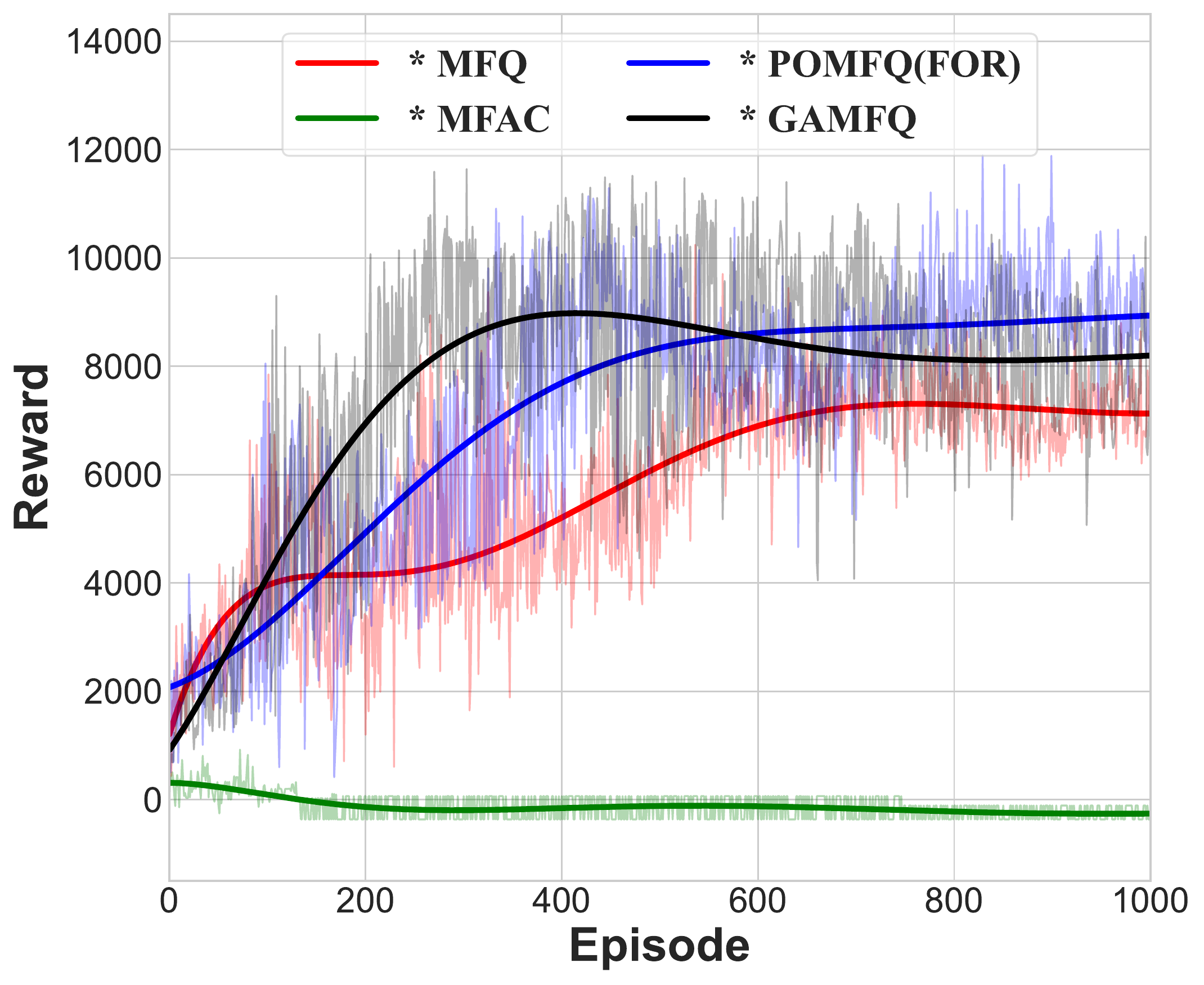}
				\end{minipage}
			}
			\subfigure[Train results of Predator-Prey game.]{
				\begin{minipage}[a]{0.3205\textwidth}
					\includegraphics[width=1\textwidth]{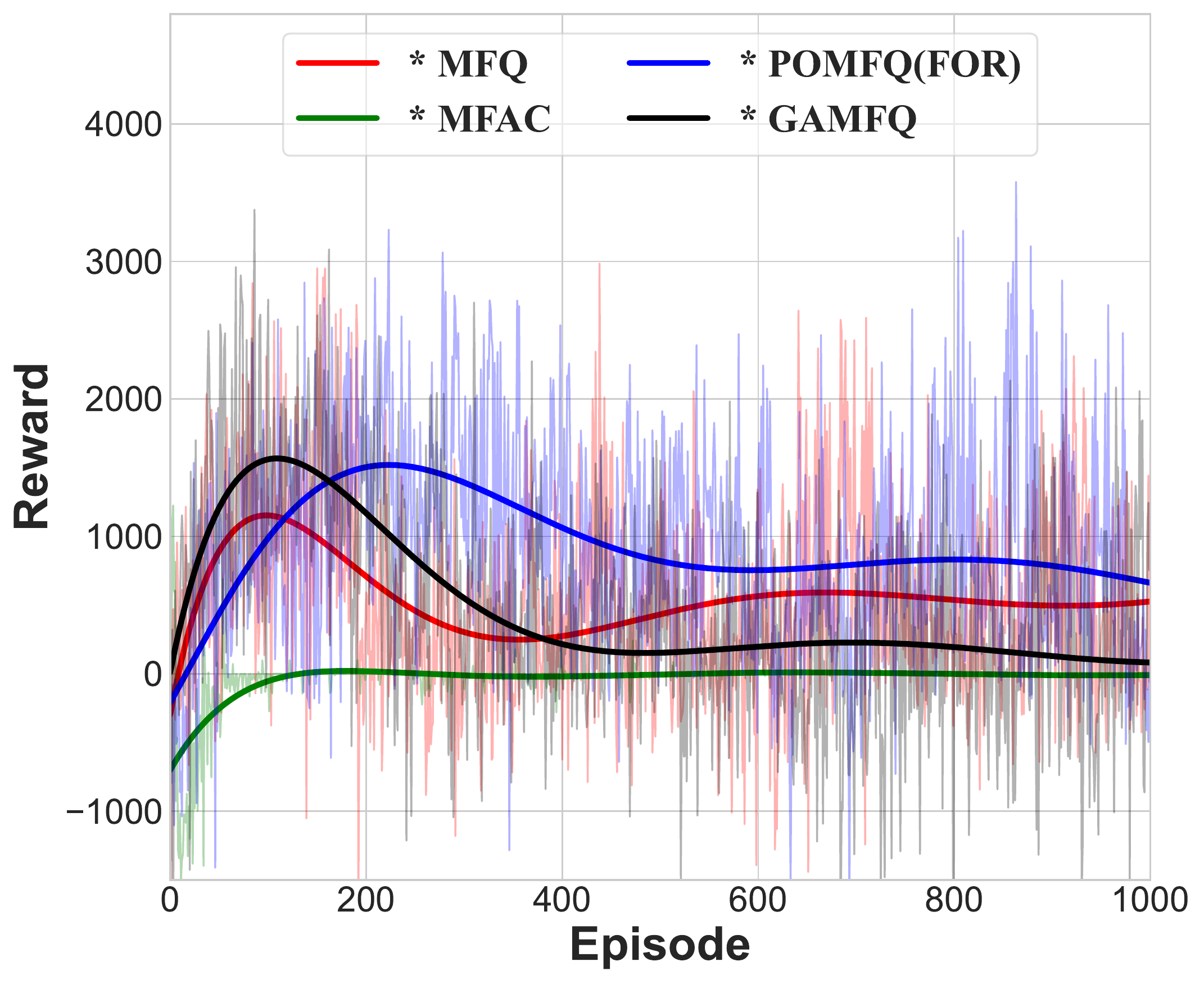}
				\end{minipage}
		}}
		\caption{Train results of three games. The reward curve for each algorithm is fitted by the least squares method.}
		\label{fig3}
	\end{figure}

	Figure \ref{fig3} shows how the reward changes as the number of iterations increases during training. We plot the reward changes for the four algorithms in different game environments during the first 1000 iterations. Since each algorithm is self-training which results in a large change in the reward of the algorithm, we use the least squares method to fit the reward change graph. In Figure \ref{fig3}, the solid black line represents the reward change graph of the GAMFQ algorithm. From Figure \ref{fig3} (a), (b) and (c), it can be seen that the reward of the GAMFQ algorithm can increase rapidly, indicating that the GAMFQ algorithm can converge rapidly in the early stage, and the convergence performance is better than the other three algorithms.

	\subsection{Elo Calculation}
	We use ELO Score \cite{Jaderberg2018HumanlevelPI} to evaluate the performance of the two groups of agents, the advantage of which is that it takes into account the strength gap between the opponents themselves. ELO ratings are commonly used in chess to evaluate one-on-one situations, and this approach can similarly be extended to N-versus-N situations. For the algorithm proposed in the paper, we record the total rewards of the two teams of agents during each algorithm confrontation, which are $R_1$ and $R_2$, respectively. Then the expected win rates of the two groups of agents are:
	\begin{eqnarray}
		{E_1} = \frac{1}{{1 + {{10}^{({R_2} - {R_1})/400}}}},{E_2} = \frac{1}{{1 + {{10}^{({R_1} - {R_2})/400}}}}
		\label{eq24}
	\end{eqnarray}
	where ${E_1} + {E_2} = 1$. By analyzing the actual and predicted winning rates of the two groups of agents, the new ELO score of each team after the game ends can be obtained:
	\begin{eqnarray}
		{R_1}^\prime  = {R_1} + K({S_1} - {E_1}),{R_2}^\prime  = {R_2} + K({S_2} - {E_2})
		\label{eq25}
	\end{eqnarray}
	where ${R_1}$ represents the actual winning or losing value, 1 means the team wins, 0.5 means the two teams are tied, and 0 means the team loses. $K$ is represented as a floating coefficient. To create a gap between agents, we set $K$ to 32. For each match, we faced off 500 times and calculated the average ELO value for all matches. 
	
	\begin{center}
		\begin{table}[!h]%
			\centering
			\caption{The ELO Score of four algorithms in Multibattle environment.\label{tab2}}%
			\begin{tabular*}{500pt}{@{\extracolsep\fill}lcccc@{\extracolsep\fill}}
				\toprule
				\textbf{Task} & \textbf{Algorithm1}  & \textbf{Algorithm2} & \textbf{ELO Score1}  & \textbf{ELO Score2} \\
				\midrule
				\multirow{2}{*}{GAMFQ vs POMFQ(FOR)} & GAMFQ-1& POMFQ(FOR)-2 & \textbf{3579} & 820   \\
				& GAMFQ-2 &  POMFQ(FOR)-1 &   2696 & \textbf{2838}  \\
				\midrule
				\multirow{2}{*}{GAMFQ vs MFQ}& GAMFQ-1 & MFQ-2 &    \textbf{2098} & 1508 \\
				& GAMFQ-2 & MFQ-1 &    \textbf{2535} & 1695  \\
				\midrule
				\multirow{2}{*}{GAMFQ vs MFAC}& GAMFQ-1 & MFAC-2 &   \textbf{1350} & -49  \\
				& GAMFQ-2 & MFAC-1 &    -856 & \textbf{-78} \\
				\midrule
				\multirow{2}{*}{POMFQ(FOR) vs MFQ}& POMFQ(FOR)-1& MFQ-2 & \textbf{3145} & 2577 \\
				& POMFQ(FOR)-2 & MFQ-1 &  2569 & \textbf{2857}\\
				\midrule
				\multirow{2}{*}{POMFQ(FOR) vs MFAC}& POMFQ(FOR)-1 & MFAC-2 &  -205 & \textbf{-64} \\
				& POMFQ(FOR)-2 & MFAC-1 & \textbf{826} & -42 \\
				\midrule
				\multirow{2}{*}{MFQ vs MFAC}& MFQ-1 & MFAC-2 &  -142 & \textbf{-49} \\		
				& MFQ-2 & MFAC-1 &  \textbf{610} &-46 \\ 
				\bottomrule
			\end{tabular*}
		\end{table}
	\end{center}
	
	\begin{center}
		\begin{table}[!h]%
			\centering
			\caption{The ELO Score of four algorithms in Battle-Gathering environment.\label{tab3}}%
			\begin{tabular*}{500pt}{@{\extracolsep\fill}lcccc@{\extracolsep\fill}}
				\toprule
				\textbf{Task} & \textbf{Algorithm1}  & \textbf{Algorithm2} & \textbf{ELO Score1}  & \textbf{ELO Score2} \\
				\midrule
				\multirow{2}{*}{GAMFQ vs POMFQ(FOR)} & GAMFQ-1& POMFQ(FOR)-2 & 7770 & \textbf{8931}   \\
				& GAMFQ-2 &  POMFQ(FOR)-1 &   8293 & \textbf{9310} \\
				\midrule
				\multirow{2}{*}{GAMFQ vs MFQ}& GAMFQ-1 & MFQ-2 & 6374    & \textbf{10870} \\
				& GAMFQ-2 & MFQ-1 &   \textbf{8510}   & 8313 \\
				\midrule
				\multirow{2}{*}{GAMFQ vs MFAC}& GAMFQ-1 & MFAC-2 & \textbf{5525}  & 10   \\
				& GAMFQ-2 & MFAC-1 & \textbf{10751}  &  -31 \\
				\midrule
				\multirow{2}{*}{POMFQ(FOR) vs MFQ}& POMFQ(FOR)-1& MFQ-2 &  8526 &  \textbf{8760} \\
				& POMFQ(FOR)-2 & MFQ-1 & \textbf{8632}  &  8227\\
				\midrule
				\multirow{2}{*}{POMFQ(FOR) vs MFAC}& POMFQ(FOR)-1 & MFAC-2 & \textbf{12722}  & 0 \\
				& POMFQ(FOR)-2 & MFAC-1 &\textbf{12171 } &   -88 \\
				\midrule
				\multirow{2}{*}{MFQ vs MFAC}& MFQ-1 & MFAC-2 & \textbf{12649}  & 49 \\		
				& MFQ-2 & MFAC-1 & \textbf{13788}  &  -48  \\ 
				\bottomrule
			\end{tabular*}
		\end{table}
	\end{center}
	Table \ref{tab2}, \ref{tab3}, \ref{tab4} shows the ELO scores of the four algorithms on the three tasks. It can be seen from Table \ref{tab2} that in Multibattle environment, the GAMFQ algorithm has the highest ELO score of 3579, which is significantly better than the other three algorithms. As shown in Table \ref{tab3}, in Battle-Gathering environment, the ELO score of the MFQ algorithm is the highest, and the ELO score of the GAMFQ algorithm is average. This is because the collection environment contains food, and some algorithms tend to eat food to get rewards quickly, rather than attacking enemy agents. However, the final game winning or losing decision is made by comparing the number of remaining agents between the two teams of agents. As shown in Table \ref{tab4}, in Predator-Prey environment, the ELO score of the GAMFQ algorithm has the highest ELO score of 860, which is significantly better than the other three algorithms. From the experimental results in the three environments, we can summarize that ELO score of the GAMFQ algorithm is better than other three algorithms, showing better performance.
	\begin{center}
		\begin{table}[!h]%
			\centering
			\caption{The ELO Score of four algorithms in Predator-Prey environment.\label{tab4}}%
			\begin{tabular*}{500pt}{@{\extracolsep\fill}lcccc@{\extracolsep\fill}}
				\toprule
				\textbf{Task} & \textbf{Algorithm1}  & \textbf{Algorithm2} & \textbf{ELO Score1}  & \textbf{ELO Score2} \\
				\midrule
				\multirow{2}{*}{GAMFQ vs POMFQ(FOR)} & GAMFQ-1& POMFQ(FOR)-2 & \textbf{421}& -32   \\
				& GAMFQ-2 &  POMFQ(FOR)-1 &   \textbf{16}& 7 \\
				\midrule
				\multirow{2}{*}{GAMFQ vs MFQ}& GAMFQ-1 & MFQ-2 &    \textbf{714} & -27\\
				& GAMFQ-2 & MFQ-1  &   \textbf{-15}  & -94 \\
				\midrule
				\multirow{2}{*}{GAMFQ vs MFAC}& GAMFQ-1 & MFAC-2 & \textbf{ 860}   &  -28  \\
				& GAMFQ-2 & MFAC-1 & 16  & 16 \\
				\midrule
				\multirow{2}{*}{POMFQ(FOR) vs MFQ}& POMFQ(FOR)-1& MFQ-2 &  \textbf{66} &  18 \\
				& POMFQ(FOR)-2 & MFQ-1 &  13 & \textbf{24}\\
				\midrule
				\multirow{2}{*}{POMFQ(FOR) vs MFAC}& POMFQ(FOR)-1 & MFAC-2 &  \textbf{16}  &  -16 \\
				& POMFQ(FOR)-2 & MFAC-1 & \textbf{47}&   16 \\
				\midrule
				\multirow{2}{*}{MFQ vs MFAC}& MFQ-1 & MFAC-2 & \textbf{16}  &  -16 \\		
				& MFQ-2 & MFAC-1 &   \textbf{174} &  17  \\ 
				\bottomrule
			\end{tabular*}
		\end{table}
	\end{center}

	\subsection{Results}
	
	\begin{figure}[htbp]
		\centering
		\subfigure[Faceoff results of Multibattle game. ]{
			\begin{minipage}[a]{0.3205\textwidth}
				\includegraphics[width=1\textwidth]{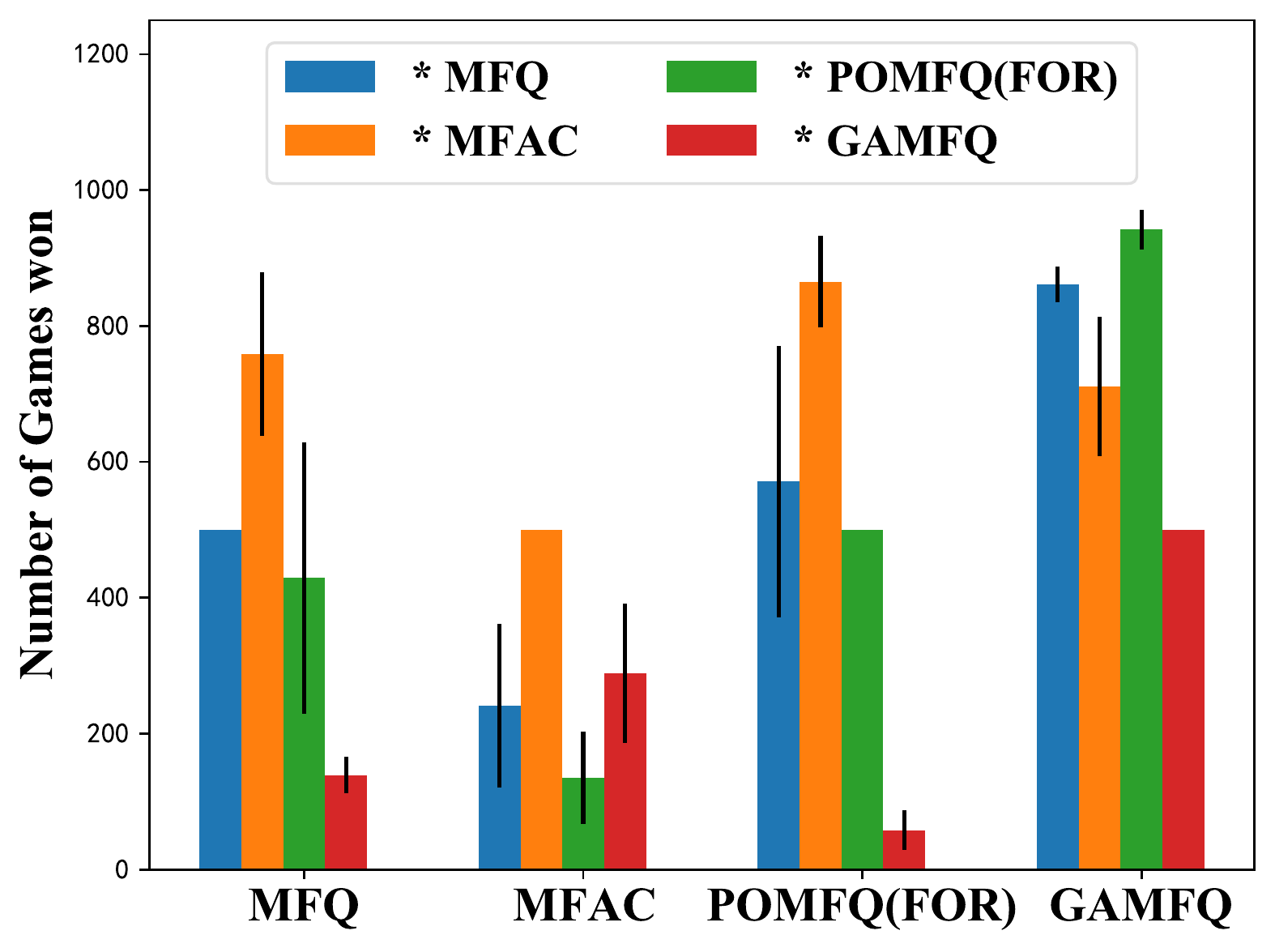}
			\end{minipage}
		}
		\subfigure[Faceoff results of Battle-Gathering game.]{
			\begin{minipage}[a]{0.3205\textwidth}
				\includegraphics[width=1\textwidth]{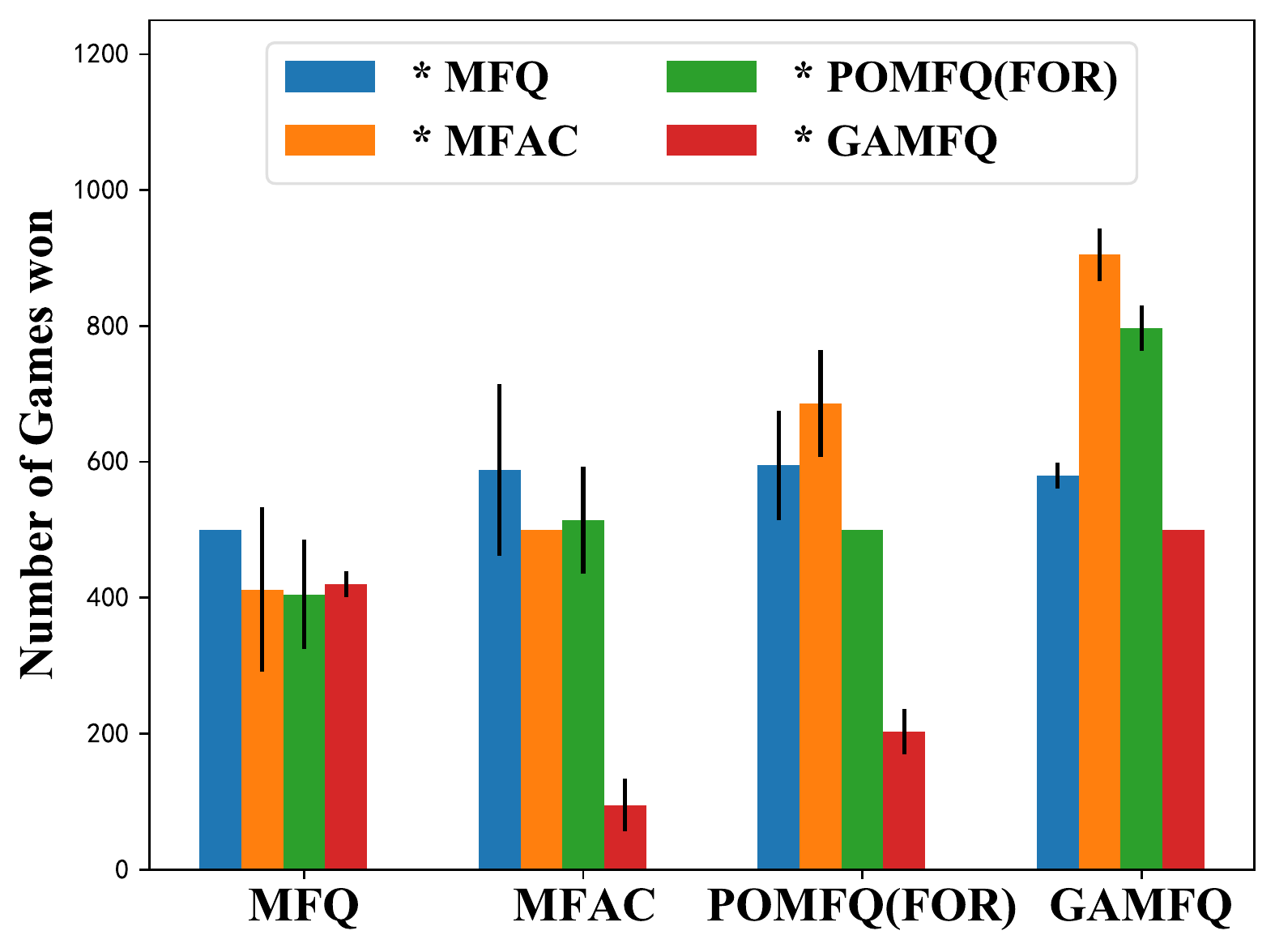}
			\end{minipage}
		}
		\subfigure[Faceoff results of Predator-Prey game.]{
			\begin{minipage}[a]{0.3205\textwidth}
				\includegraphics[width=1\textwidth]{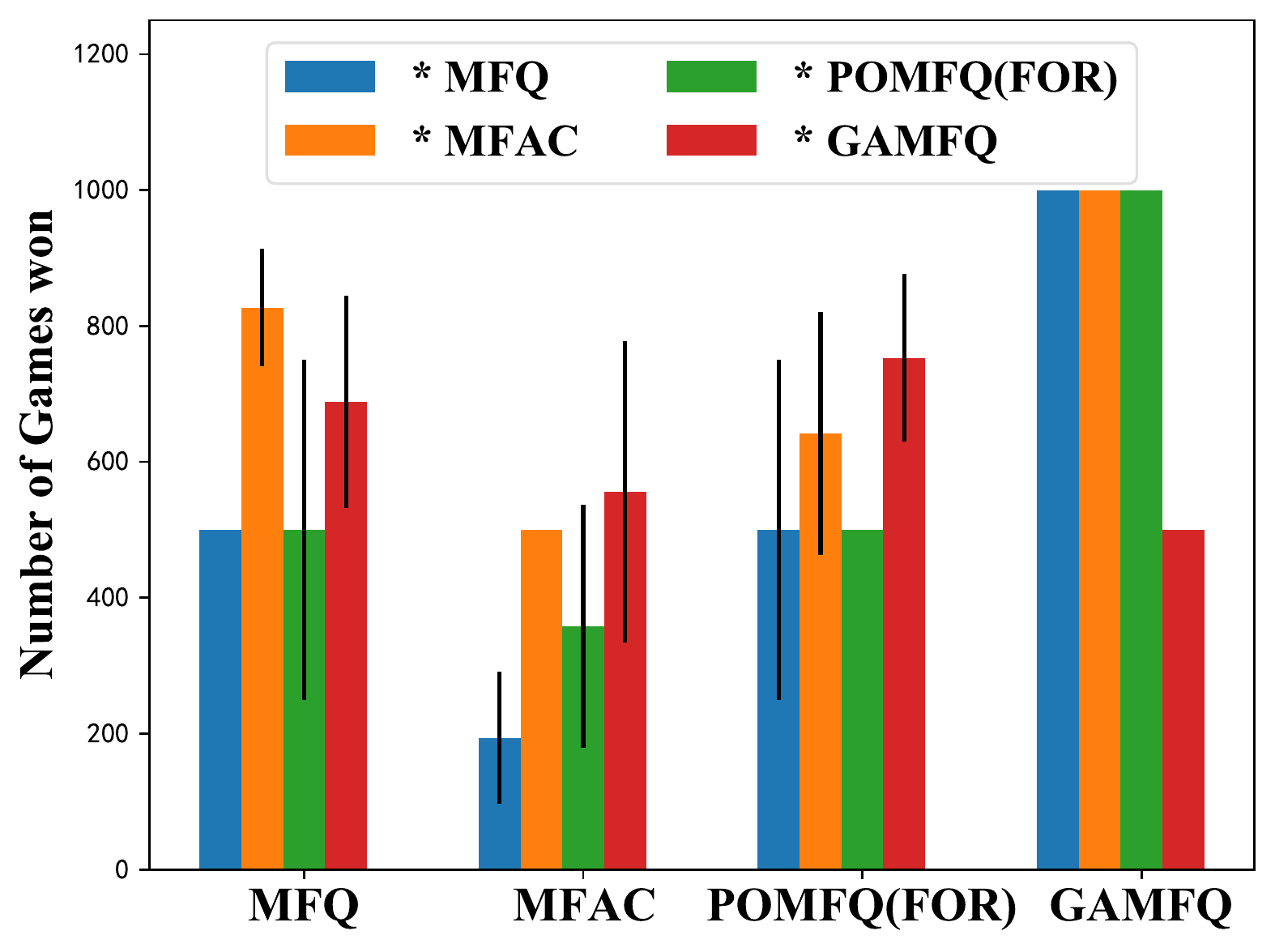}
			\end{minipage}
		}
		\caption{Faceoff results of three games. The * in the legend indicates the enemy. For example, the first blue bar in the bar graph corresponding to the GAMFQ algorithm is the result of the confrontation between GAMFQ and MFQ, and we do not conduct confrontation experiments between the same algorithms.}
		\label{fig4}
	\end{figure}
	
	Figure 4 shows the face-off results of the four algorithms in the three tasks. Figure \ref{fig4}(a) shows the faceoff results of Multibattle game. The different colored bars for each algorithm represent the results of an algorithm versus others. We do not conduct adversarial experiments between the same algorithms because we consider that the adversarial properties of the same algorithms are equal. The vertical lines in the bar graph represent the standard deviation of wins for groups A and B over 1,000 face-offs. Figure \ref{fig4}(a) shows GAMFQ against three other algorithms, all with a win rate above 0.7.

	Figure \ref{fig4}(b) shows the faceoff results of Battle-Gathering game. In addition to getting rewards for killing enemies, agents can also get rewards from food. It can be seen that MFQ loses to all other algorithms, MFAC and POMFQ (FOR) perform in general, and our GAMFQ is clearly ahead of other algorithms.
	
	Figure \ref{fig4}(c) shows thwe faceoff results of Predator-Prey game.The standard deviation of this game is significantly higher than the previous two games, due to the fact that both groups A and B are trying to beat each other in the environment. It can be seen that the GAMFQ algorithm is significantly better than other three algorithms, reaching a winning rate of 1.0.

	Experiments in the above three multi-agent combat environments show that GAMFQ can show good performance over  MFQ, MFAC and POMFQ(FOR).
	\subsection{Visualization}
	\begin{figure}[!h]
		\begin{center}	
			\includegraphics[width=0.7\textwidth]{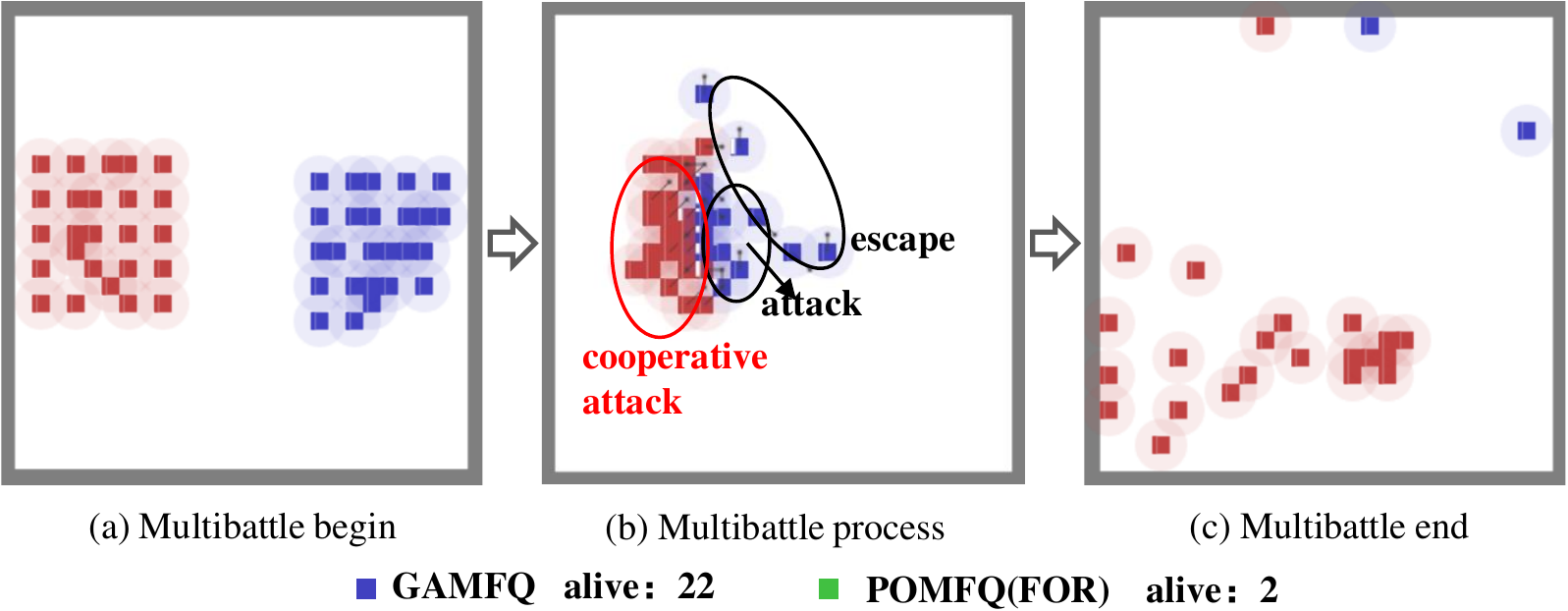}		
		\end{center}	
		\caption{Visualization of the standoff between GAMFQ and POMFQ (FOR) in a Multibattle game.}	
		\label{fig5}
	\end{figure}
	\begin{figure}[!h]
		\begin{center}	
			\includegraphics[width=0.7\textwidth]{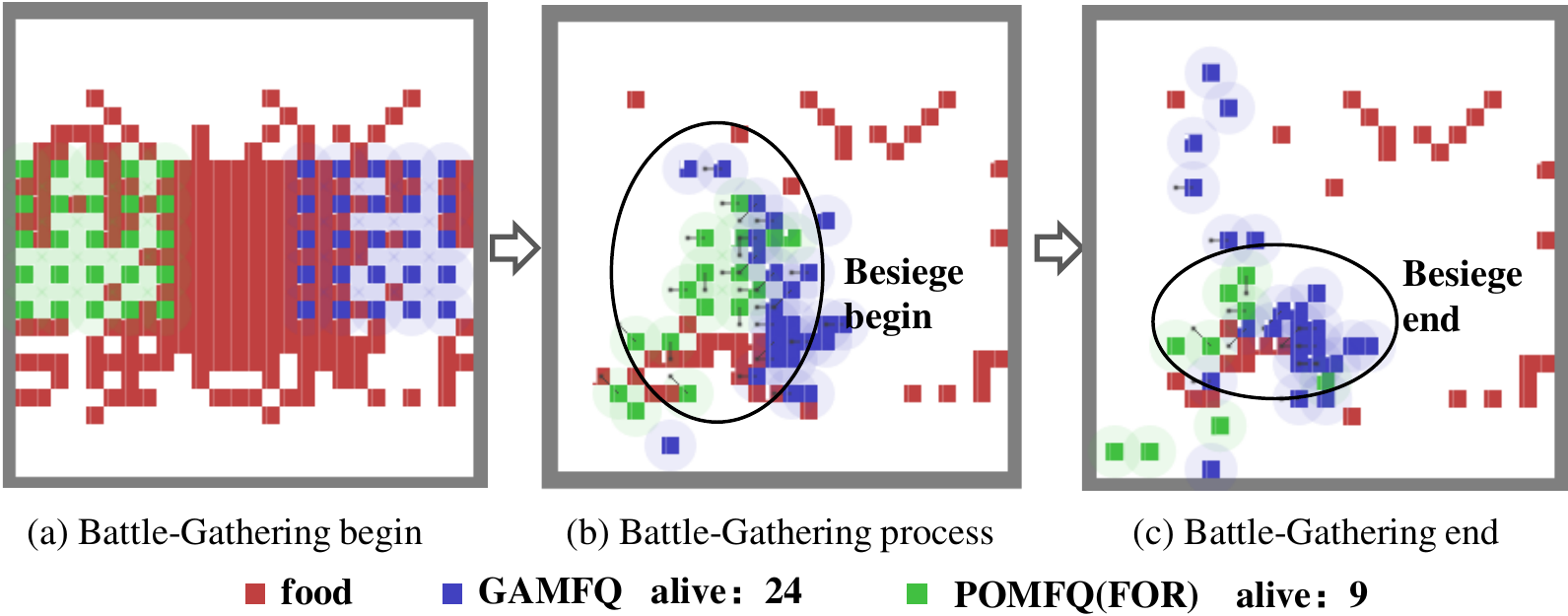}		
		\end{center}	
		\caption{Visualization of the standoff between GAMFQ and POMFQ (FOR) in a Battle-Gathering game.}	
		\label{fig6}
	\end{figure}
	\begin{figure}[!h]
		\begin{center}	
			\includegraphics[width=0.7\textwidth]{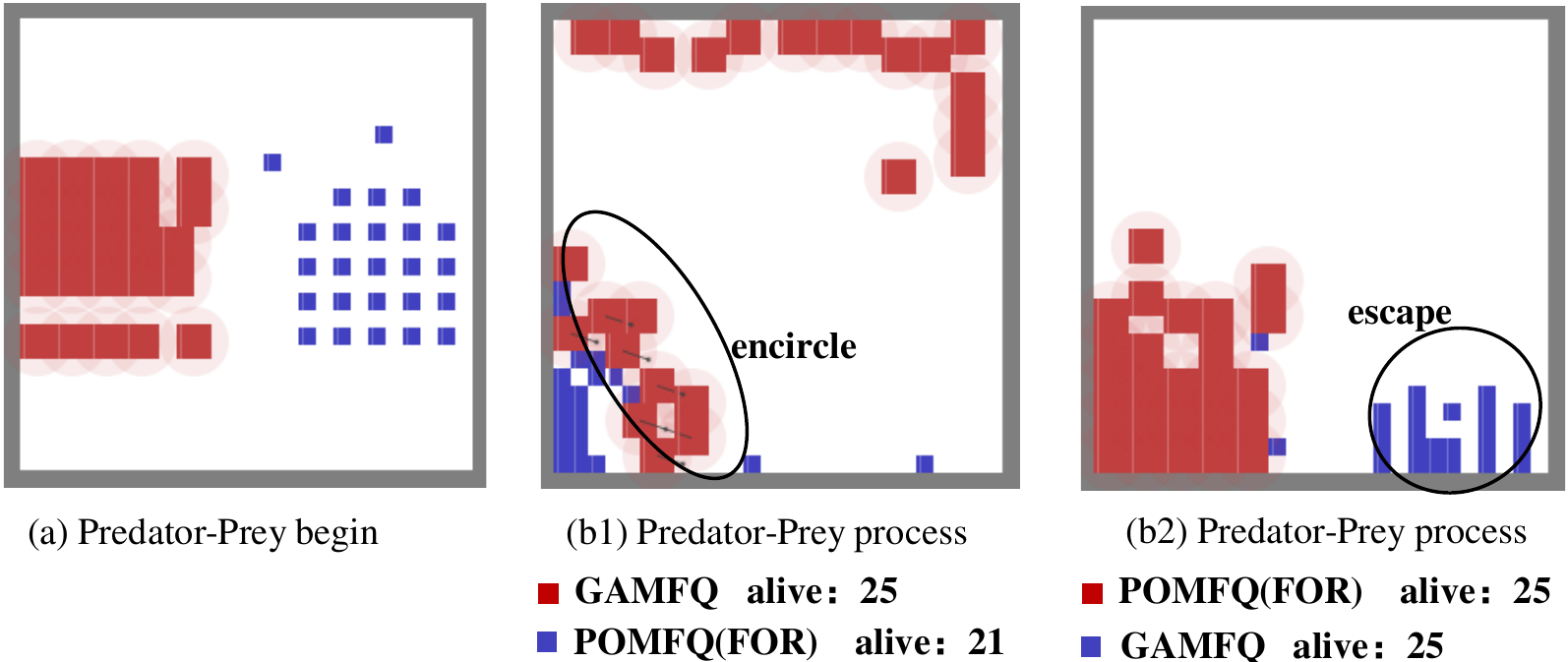}		
		\end{center}	
		\caption{Visualization of the standoff between GAMFQ and POMFQ (FOR) in a Predator-Prey game.}	
		\label{fig7}
	\end{figure}

	To visualize the effectiveness of the GAMFQ algorithm, we visualize the confrontation between GAMFQ and POMFQ (FOR) in a Multibattle environment, as shown in Figure \ref{fig5}, where the red side is GMAFQ and the blue side is POMFQ (FOR). It can be seen from the confrontation process that for the GAMFQ algorithm, when an agent decides to attack, the surrounding agents will also decide to attack under its influence, forming a good cooperation mechanism. On the contrary, for the POMFQ (FOR) algorithm, some blue-side agents are chosen to attack, some are chosen to escape, and no common fighting mechanism was formed. Similarly, in the Battle-Gathering environment of Figure \ref{fig6}, GAMFQ can learn the surrounding mechanism well. In the Predator-Prey environment of Figure \ref{fig7}, when GAMFQ acts as a predator, the technique of surrounding the prey POMFQ (FOR) can be learned. On the contrary, when POMFQ (FOR) acted as a predator, it failed to catch the prey GMAFQ.

	\subsection{Ablation study}
	
	\begin{figure}[!h]
		\centerline{\includegraphics[width=7cm]{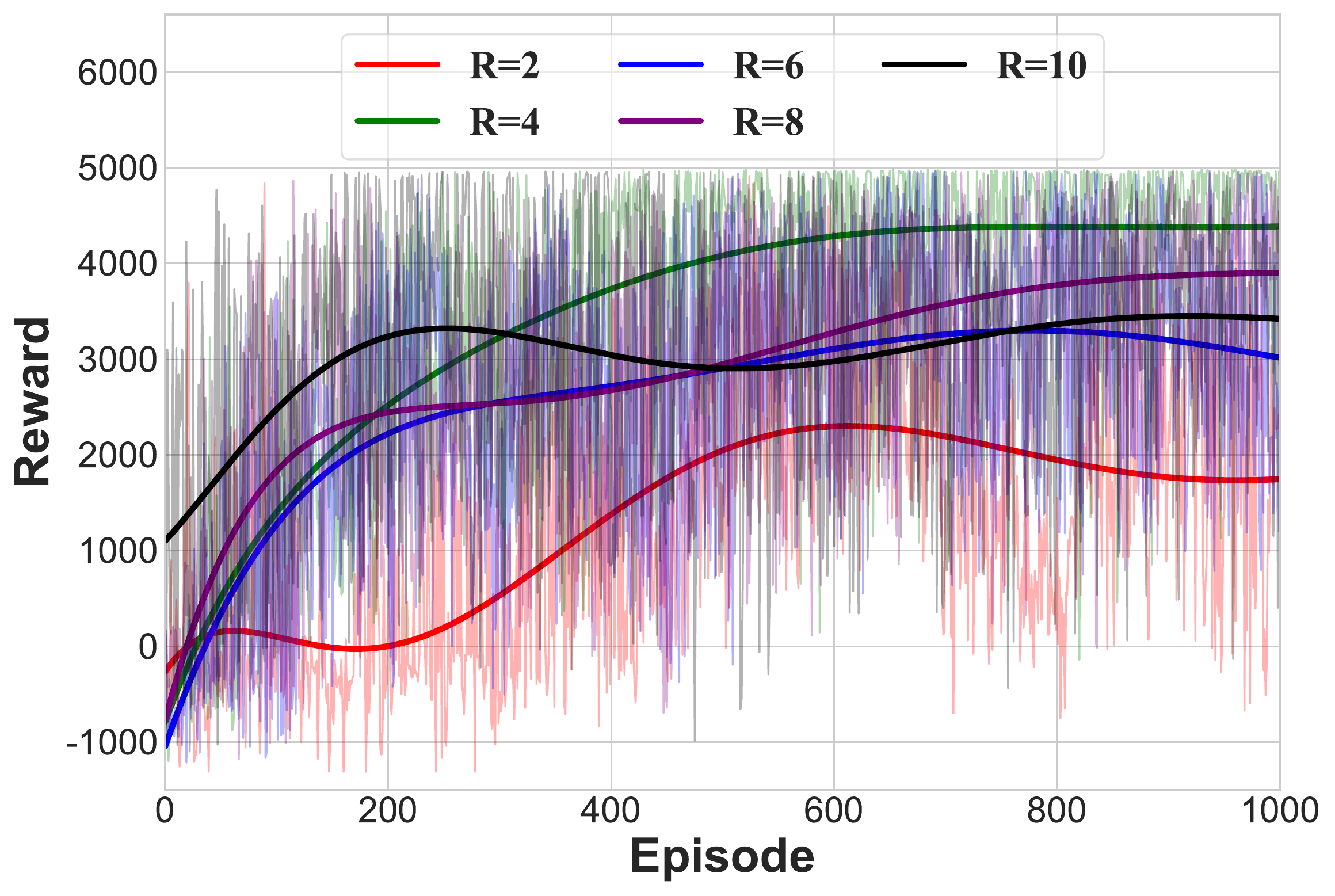}}
		\caption{Ablation study. R represents the observation radius of the agent.\label{fig8}}
	\end{figure}
	
	Figure \ref{fig8} is an ablation study that investigates the performance of the GAMFQ algorithm for different observation radius in a Multibattle environment. where the solid line represents the least squares fit of the reward change. It can be seen from the figure that when the number of training is small, the performance of the algorithm is better as the observation distance increases. But with the increase of training times, when R=4, the performance of the algorithm is the best, so the appropriate observation distance can achieve better performance. What is more important in this paper is to explore the effect of the ratio of observable distance to the number of agents on the performance of the algorithm, so there is no experiment with more agents.
	
	\section{Conclusion}
	
	In this paper, we proposed a new multi-agent reinforcement learning algorithm, Graph Attention-based Partially Observable Mean Reinforcement Learning (GAMFQ), to address the problem of large-scale partially observable multi-agent environments. Although existing methods are close to Nash equilibrium, they do not take into account the direct correlation of agents. Based on the correlation between agents, GAMFQ uses a Graph-Attention module to describe how each agent is affected by the actions of other agents at each time step. Experimental results on three challenging tasks in the MAgents framework illustrate that, our proposed method outperforms baselines in all these games and outperforms the state-of-the-art partially observable mean-field reinforcement learning algorithms. In the future, we will further explore the correlation between agents to extend to more common cooperation scenarios.

	%\backmatter
	
	%\section*{Acknowledgments}
	%This work is supported by xxxx, China (project no.xxxx).
	
	%\subsection*{Author contributions}
	%\textbf{Min Yang:} Writing – original draft.
	%\textbf{Guanjun Liu:} Conceptualization; writing – review and editing.
	%\textbf{Ziyuan Zhou:} Formal analysis.
	
	%\subsection*{Financial disclosure}
	%
	%None reported.
	
	\subsection*{Conflict of interest}
	
	The authors declare no potential conflict of interests.
	
	\subsection*{Article Description}	
	
	The expanded version of this article is published in Drones 2023, 7(7), 476, with a DOI of https://doi.org/10.3390/drones7070476.
	
	\nocite{*}% Show all bib entries - both cited and uncited; comment this line to view only cited bib entries;
	
	\bibliography{wileyNJD-AMS}%
	
	%\section*{Author Biography}
	%
	%\begin{biography}{\includegraphics[width=60pt,height=70pt,draft]{empty}}{\textbf{Author Name.} This is sample author biography text this is sample author biography text this is sample author biography text this is sample author biography text this is sample author biography text this is sample author biography text this is sample author biography text this is sample author biography text this is sample author biography text this is sample author biography text this is sample author biography text this is sample author biography text this is sample author biography text this is sample author biography text this is sample author biography text this is sample author biography text this is sample author biography text this is sample author biography text this is sample author biography text this is sample author biography text this is sample author biography text.}
	%\end{biography}
	
\end{document}